\documentclass[a4paper,fleqn]{cas-dc}

\usepackage[round]{natbib}        
\let\cite\citep                   

\usepackage{subcaption}

\def\tsc#1{\csdef{#1}{\textsc{\lowercase{#1}}\xspace}}
\tsc{WGM}
\tsc{QE}
\tsc{EP}
\tsc{PMS}
\tsc{BEC}
\tsc{DE}

\begin{document}
\let\WriteBookmarks\relax
\def\floatpagepagefraction{1}
\def\textpagefraction{.001}

\shorttitle{}

\shortauthors{Yanchen Guan et~al.}

\title [mode = title]{Domain-Enhanced Dual-Branch Model for Efficient and Interpretable Accident Anticipation}                      


\author[1,2]{Yanchen Guan}\credit{Conceptualization of this study, Methodology, Experiment, Writing}
\author[1,3]{Haicheng Liao}\credit{Conceptualization of this study, Writing}
\author[1,2]{Chengyue Wang}\credit{Writing}
\author[1,3]{Bonan Wang}\credit{Experiment}
\author[1,2]{Jiaxun Zhang}\credit{Experiment}
\author[4]{Jia Hu}\credit{Conceptualization of this study}
\author[1,2,3]{Zhenning Li}[orcid=0000-0002-0877-6829]\cormark[1]\credit{Conceptualization of this study,Writing}\ead{zhenningli@um.edu.mo}


\cortext[cor1]{Corresponding author}

\affiliation[1]{organization={State Key Laboratory of Internet of Things for Smart City},
    addressline={University of Macau}, 
    city={Macau SAR},
    postcode={999078}, 
    country={China}}
\affiliation[2]{organization={Department of Civil Engineering},
    addressline={University of Macau}, 
    city={Macau SAR},
    postcode={999078}, 
    country={China}}
\affiliation[3]{organization={Department of Computer and Information Science},
    addressline={University of Macau}, 
    city={Macau SAR},
    postcode={999078}, 
    country={China}}
\affiliation[4]{organization={College of Transportation Engineering},
    addressline={Tongji University}, 
    city={Shanghai},
    postcode={200092}, 
    country={China}}

\begin{abstract}
Developing precise and computationally efficient traffic accident anticipation system is crucial for contemporary autonomous driving technologies, enabling timely intervention and loss prevention. In this paper, we propose an accident anticipation framework employing a dual-branch architecture that effectively integrates visual information from dashcam videos with structured textual data derived from accident reports. Furthermore, we introduce a feature aggregation method that facilitates seamless integration of multimodal inputs through large models (GPT-4o, Long-CLIP), complemented by targeted prompt engineering strategies to produce actionable feedback and standardized accident archives. Comprehensive evaluations conducted on benchmark datasets (DAD, CCD, and A3D) validate the superior predictive accuracy, enhanced responsiveness, reduced computational overhead, and improved interpretability of our approach, thus establishing a new benchmark for state-of-the-art performance in traffic accident anticipation.
\end{abstract}




\maketitle

\section{Introduction}
The vision of fully autonomous vehicles (AVs) promises to significantly enhance road safety, yet the persistent reality of traffic accidents, which remain a leading cause of fatalities worldwide, underscores the challenges that lie ahead \cite{ahmed2023road}. As AVs become more integrated into our urban landscapes, the ability to predict and prevent accidents in real-time becomes crucial \cite{liao2024when, abdel2024matched}. Accident anticipation can predict potential traffic accidents that are about to occur within the field of view through dashcam video input, provide early warnings to drivers or autonomous driving systems, thereby enabling timely emergency responses and ensuring ego vehicle can avoid possible traffic accidents or the impact of traffic accidents.\cite{fang2023vision} However, current methods for accident anticipation often fall short due to their inability to fully address the complexity of dynamic traffic environments \cite{chen2022review}. Conventional systems typically focus on detecting and modeling interactions among individual traffic agents using techniques such as object detection, depth estimation, and optical flow analysis, with these features being processed by graph convolutional networks (GCNs) \cite{zhang2019graph,liao2024real} to establish spatial-temporal relationships. While these approaches are effective in controlled settings, they face significant challenges when applied to real-world, dynamic traffic scenarios:

Firstly, traditional methods rely heavily on object detection and segmentation to track traffic agents, but this approach becomes increasingly impractical in multi-agent scenarios \cite{mishra2023sensing, le2020attention}. The need to process and analyze numerous objects simultaneously creates substantial computational burdens, limiting scalability and real-time performance. Secondly, by concentrating on individual entities, traditional models often miss the broader contextual information necessary for accurate accident anticipation \cite{kumar2020iot}. This narrow focus can result in a fragmented understanding of traffic scenes, hindering the ability to predict complex, real-world scenarios where context and subtle cues are critical. Finally, deep learning models \cite{fang2022traffic, karim2022dynamic}, commonly used in these systems, often suffer from a lack of transparency, making their predictions difficult to interpret. This opacity poses significant challenges in safety-critical applications, where understanding the reasoning behind forecasts is crucial for ensuring reliable decision-making.

\begin{figure}[t]
\centering  
\includegraphics[width=0.5\textwidth]{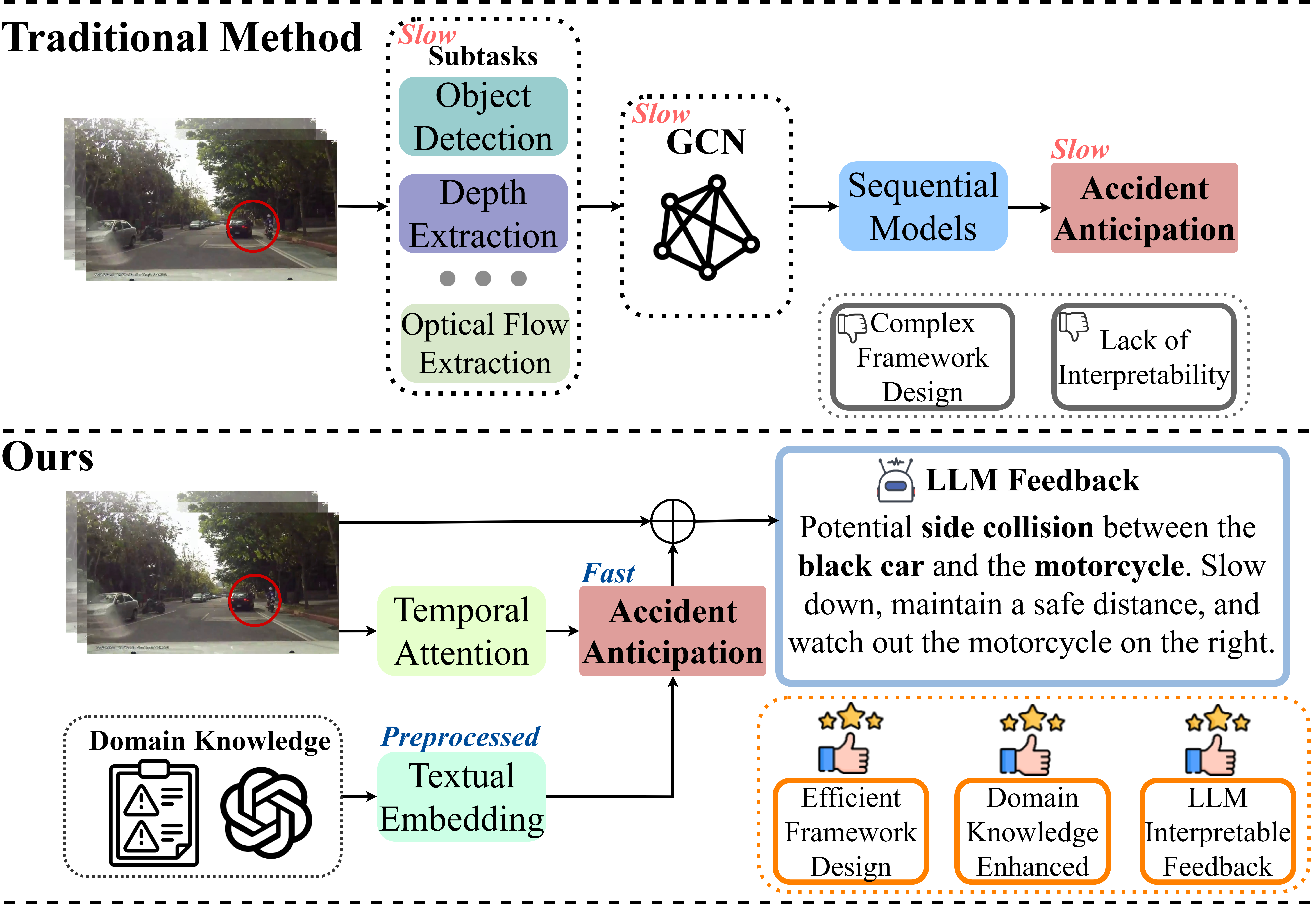}

\caption{{Comparison of traditional and proposed accident anticipation models. The traditional approach (top) relies on object detection, depth estimation, and optical flow processed through a Graph Convolutional Network for accident anticipation. In contrast, our proposed method (bottom) integrates domain knowledge and a Large Language Model (GPT-4o) to enhance interpretability and provide more context-aware feedback.}}
\label{Fig.1}
\end{figure}

We posit that driving is fundamentally guided by domain knowledge—socially established norms and rules, such as the interactions between drivers and the safe driving behaviors mandated by traffic regulations. This domain knowledge is rooted in patterns derived from extensive human driving experience and daily observations, rather than being an innate ability of human drivers. Inspired by this insight, we redefine accident anticipation as a task that integrates both knowledge-driven and data-driven approaches. We advocate for the injection of domain knowledge into neural networks to guide the model’s learning process according to established paradigms, thereby maintaining realism and enhancing generalization.
In response to these challenges, we propose a novel framework that shifts the focus from an agent-centric approach to a scene-centric, domain-informed methodology. By integrating large language models (LLMs) fine-tuned with structured accident reports, our model leverages domain knowledge to enhance predictive capabilities while overcoming the limitations of traditional methods.

Overall, our contributions are as follows:
\begin{itemize}

\item We introduce a novel dual-branch multimodal learning architecture that effectively aligns visual and linguistic features, enabling a comprehensive understanding of complex traffic scenes. One branch extracts temporal attention from visual data, while the other leverages structured accident reports as prior knowledge to enhance predictive accuracy.

    \item Unlike traditional methods that rely on individual object detection, our model prioritizes scene-level feature extraction, significantly reducing computational complexity. This efficiency enables real-time operation, overcoming bottlenecks associated with high-dimensional input processing.

    \item We propose a structured framework for leveraging and archiving accident reports within the proposed model. By integrating visual language models, our model not only utilizes historical accident reports for risk assessment but also archives new accident scenarios to refine future predictions. This self-improving mechanism enhances data utilization, reduces human annotation effort, and mitigates safety risks.

    \item Our model undergoes extensive evaluation on multiple benchmark datasets, consistently outperforming top methods in both accuracy and mean time-to-accident metrics. Additionally, the structured feedback generated by LLMs improves interpretability and transparency, making our approach well-suited for safety-critical autonomous driving applications.

\end{itemize}

This paper is organized as follows: In Section \ref{Related Work}, we review the relevant literature in the field.  Section \ref{Method} delves into the intricacies of accident anticipation and presents our proposed model.  Section \ref{Experiments} evaluates the performance of our model on the DAD, CCD, A3D datasets, including anticipation accuracy, computational efficiency, and interpretability. Finally, in Section \ref{Conclusion}, we offer concluding thoughts.

\section{Related Work}
\label{Related Work}
\textit{Accident Anticipation.}
Accident anticipation is an evolving field that has garnered increasing attention due to its potential to enhance road safety by predicting accidents before they occur. Early work by \citet{chan2017anticipating} introduced the concept of traffic accident anticipation, distinguishing it from accident detection, which focuses on post-incident identification. They proposed a Dynamic Spatial Attention (DSA) framework, highlighting the importance of spatial attention in identifying high-risk scenarios, and laying a foundation for subsequent research. 

In light of this work, a major research direction focuses on modeling interactions between traffic agents. \citet{karim2022dynamic} introduced the Dynamic Spatio-Temporal Attention (DSTA) framework, integrating spatial and temporal dynamics to improve prediction accuracy. GCNs have also been employed to represent traffic scenes as graphs, capturing complex spatial-temporal relationships among traffic participants \cite{fang2023vision,zhao2019t, wang2023gsc,song2024dynamic}. Additionally, attention-based methods \cite{fang2019dada,fang2021dada,fatima2021global,li2024cognitive,bao2021drive} enhance feature extraction and improve model interpretability. However, they struggle with computational efficiency and scalability in dense scenes. 

\textit{Video Feature Extraction.}
In the field of accident anticipation, the commonly used video feature extraction methods are mainly Transformer and VGG-16 \cite{fang2023vision,vaswani2017attention,simonyan2014very}. Transformer is an attention-based sequence modeling with high computational complexity but a stronger ability to understand global information; VGG-16 is based on convolutional neural networks and focuses on efficient and real-time visual feature extraction. In addition, some studies have also tried to use models such as I3D or VIT for feature extraction, but these highly complex models consume a lot of additional computing resources and have limited effects on accident anticipation \cite{liao2024when,thakur2024graph}. The feature extraction model we used in this study is VGG-16. By maintaining the same feature extraction method, we can better prove the superiority of our method in accident anticipation compared to past studies. In addition, choosing VGG-16 is also in line with our pursuit of real-time performance and low computing resource consumption.

\textit{Domain Knowledge Integration.}
Recently, domain knowledge integration has proven valuable for enhancing time-series traffic prediction by incorporating contextual factors such as traffic patterns, signage, and environmental conditions. By incorporating domain-specific expertise, rules, or prior knowledge into machine learning models or systems, the performance, interpretability, and robustness of the models can be improved \cite{dash2022review}. Common domain knowledge integration methods include feature engineering, transfer learning and fine-tuning, physics-informed models, and hybrid expert systems, etc \cite{nie2025data}. Prior studies have explored the impact of structured knowledge in trajectory forecasting and network modeling \cite{xu2022trajectory, li2022network,da2024open,parsa2019real,dan2012multi}. However, the systematic incorporation of domain knowledge into accident anticipation remains underexplored, and our study is among the first to address this gap. 

\textit{Large Models.}
Since the concept of "artificial intelligence" was proposed in the 1950s, the development of large models has begun in its infancy. Based on the main architecture Transformer \cite{vaswani2017attention}, the introduction of BERT \cite{devlin2019bert} and GPT-1 \cite{radford2018improving} marked the rise of pre-trained large models in the field of natural language processing. Subsequently,
The emergence of visual language models (VLMs) and LLMs marks a significant improvement in the ability of deep learning models to understand visual and textual information, providing new opportunities to improve the interpretability and accuracy of accident anticipation. CLIP \cite{radford2021learning} shows the capability of aligning visual and textual data, enabling zero-shot predictions, while LLMs such as the GPT series \cite{achiam2023gpt} provide contextually rich insights crucial for safety-critical applications \cite{liao2024gpt,zhou2024gpt,wang2024accidentgpt}. 
In research related to traffic accidents, large models can not only be used for human-computer interaction, but can also be directly used for image perception and analysis, thereby performing tasks such as accident anticipation, prediction, and prevention \cite{wang2024accidentgpt,yuan2024unist}. 
Our research leverages these advancements by integrating Long-CLIP \cite{zhang2024long} for cross-modal scene understanding and GPT-4o \cite{achiam2023gpt} for generating structured, actionable risk assessments, addressing the limitations of previous models in interpretability and real-time applicability.

Overall, despite these advancements, challenges remain in computational efficiency, scalability, and interpretability of existing accident anticipation models. Existing models often fail to capture the full complexity of dynamic, multi-agent traffic environments. This study bridges these gaps by introducing a novel framework that combines scene-level analysis, domain-informed reasoning, and advanced foundation models. This approach raises the standard for predictive accuracy and transparency in accident anticipation, making autonomous systems more reliable and interpretable in real-world deployments. At the same time, we have greatly improved the real-time performance of accident prediction by optimizing the model's computational efficiency.

\section{Methodology}
\label{Method}
\begin{figure*}[t]
\centering  
\includegraphics[width=0.95\textwidth]{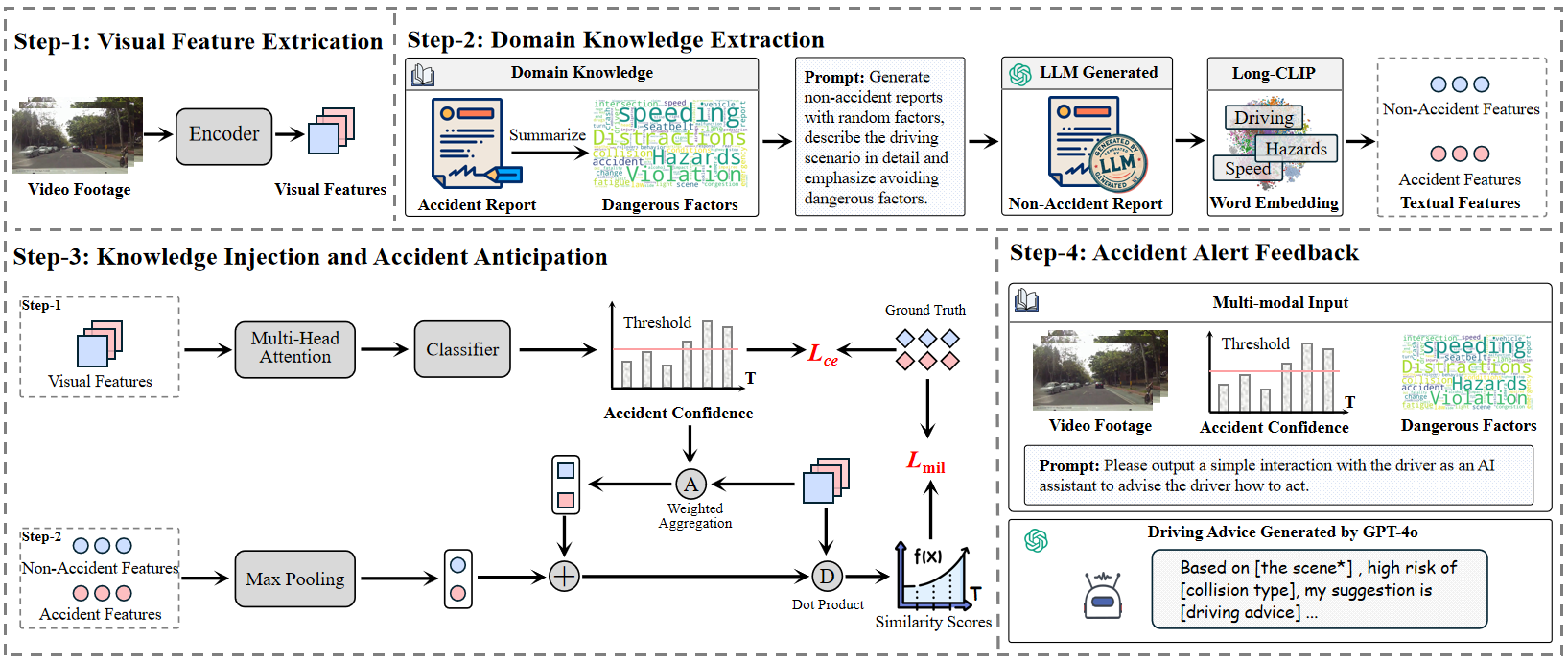}

\caption{Framework of our proposed model. It consists of four key steps, covering the process from visual and textual input to accident anticipation and alert feedback. Our proposed dual-branch framework assists visual classifiers for accident anticipation by aligning visual tokens with textual tokens in the latent space. In terms of alert feedback, through domain knowledge injection and fine-tuning of prompts based on the chain-of-thought, the LLM can better serve pre-accident scenarios based on prior knowledge.}
\label{Fig.2}
\end{figure*}

\subsection{Problem Setup}
The primary objective of this study is to predict whether an accident is likely to occur based on observations from dashcam video footage and, if so, to issue an early warning. Formally, let \( X = \{x^0, x^1, \ldots, x^T\} \) represent a sequence of observed video frames, where \( T \) is the total number of frames. The model is tasked with predicting the probability of an accident occurring at each time step \( t \), denoted as \( P = \{p^0, p^1, \ldots, p^T\} \), where \( p^t \) is the probability of an accident at time step \( t \). An accident is considered to occur if the predicted probability first exceeds a predefined threshold \( p^{\tau} \) at time step \( \bar{t} \), i.e., \( p^t \geq p^{\tau} \) where \( \bar{t} < T \).
To measure how early the model anticipates accidents, we define the Time-to-Accident (TTA) as \( \text{TTA} = \tau - \bar{t} \). Here, \( \tau \) is the actual accident occurrence time, with \( \tau = 0 \) for non-accident videos. Notably, a higher TTA value indicates the model’s capacity to predict accidents earlier. The goal of this paper is to accurately predict the probability \( P \) of accident occurrence and to maximize \( \bar{t} \) for videos that contain accidents.

\subsection{Overall Framework}
As illustrated in Fig. \ref{Fig.2}, our method consists of four key steps: Visual Feature Extraction, Domain Knowledge Extraction, Knowledge Injection and Accident Anticipation, and Accident Alert Feedback. The first step focuses on extracting frame features from the raw videos. The second step is pivotal to our approach, where we extract domain knowledge from real-world autonomous driving accident reports. Using this knowledge, we apply regularized prompts with VLMs to incrementally generate dialogues that describe corresponding non-accident reports. In the third step, these reports are then processed by a Text Encoder to produce textual features, which are compared with the frame features to calculate similarity scores for each traffic scene. Furthermore, the model utilizes these similarity scores, along with the frame features, to predict the probability of an accident occurring in each video frame. Finally, in the fourth step, the LLMs achieve the model's predictions and similarity scores to deliver real-time verbal accident alerts and prompts, thereby enhancing the safety and human-AI interaction within AD systems.

\subsection{Step-1: Visual Feature Extraction}
This step is primarily responsible for encoding the input dashcam video to obtain frame features for each video. These features capture the global semantic information of the entire traffic scene, such as lane lines, road signs, traffic lights, the overall positional relationship of vehicles, and scene dynamic changes. Using subtle scene semantic clues in visual features that affect the occurrence of accidents, such as the positional relationship between vehicles and lanes, abnormal driving trajectories, or dangerous approach movements, our model can capture clues that may be related to accidents, thereby achieving accident prediction.

Existing methods typically rely on pre-train object detectors to identify objects in the traffic scene, and their inference relies heavily on predictions from all possible candidate regions. The performance of these methods is often limited by the quality of the detected objects or the predefined anchor boxes. In real-world driving scenarios, objects involved in traffic accidents often exhibit significant variations in shape, position, and relational context. Moreover, adverse weather conditions and low-light environments further obscure the clarity of road users in visual perception systems. As a result, these detector-based approaches tend to miss critical contextual information, particularly subtle scene semantics between bounding boxes such as lane lines, road markings, and traffic signals. In addition, the inclusion of object detectors increases computational complexity, hindering accurate and real-time accident anticipation for AVs.

In contrast, our approach employs a lightweight, detector-free method, directly utilizing VGG-16 \cite{Simonyan15} to extract global contextual information rather than relying on bounding boxes generated by detectors, producing the frame features $V_f \in \mathbb{R}^{ N \times D}$, where $N$ is the number of frames, and $D$ is the dimension for hidden states.

\subsection{Step-2: Domain Knowledge Extraction}
This step aims to extract the distilled insights from real-world driving scenarios, so that in subsequent steps the accident anticipation process can be supervised and guided by extensive human driving experience. Specifically, we feed a series of accident reports, rich in traffic-related semantic information, into GPT-4 and use a unified prompt to generate corresponding non-accident reports. These reports are then mapped into a high-dimensional semantic space by a text encoder to extract critical textual features.

\begin{figure}[t]
  \subfloat [    Word Cloud]{\includegraphics[width=0.225\textwidth]{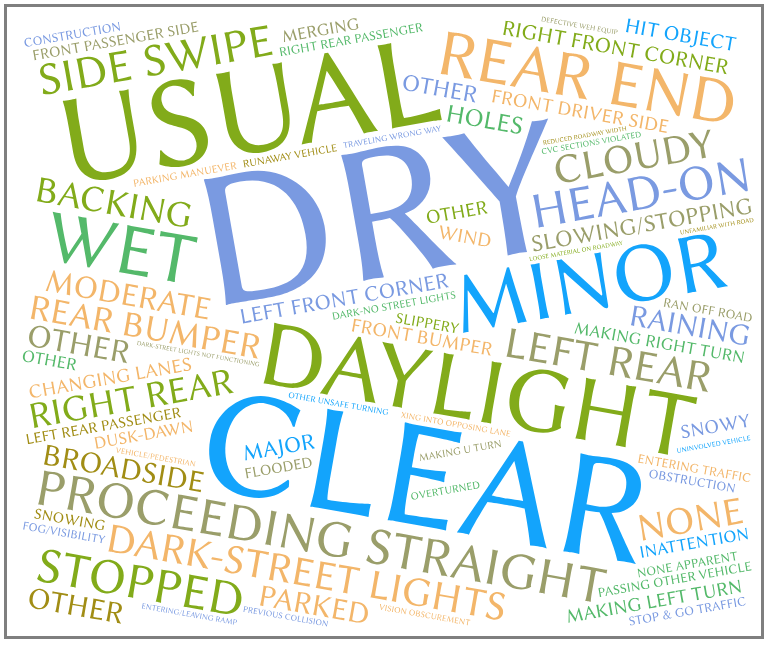}}
 \hfill 	
  \subfloat[    Distribution of Accidents]{\includegraphics[width=0.225\textwidth]{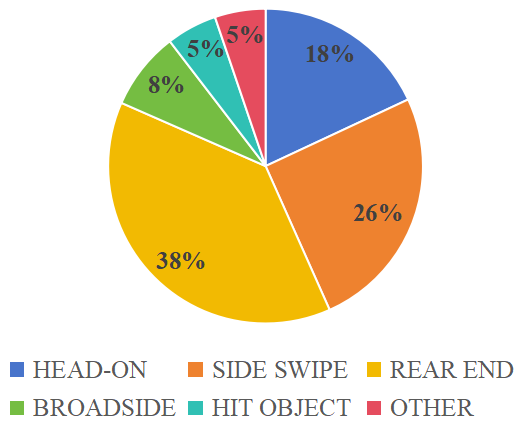}}  
\caption{ {Statistics of accident reports. (a) A word cloud generated from accident reports highlights the most frequent accident-related factors. (b) A pie chart categorizes accidents by type, with side impacts and rear-end collisions being the most common types of accidents.
}}
\label{Fig.3}
\end{figure}

\begin{figure}[t]
\centering  
\includegraphics[width=0.45\textwidth]{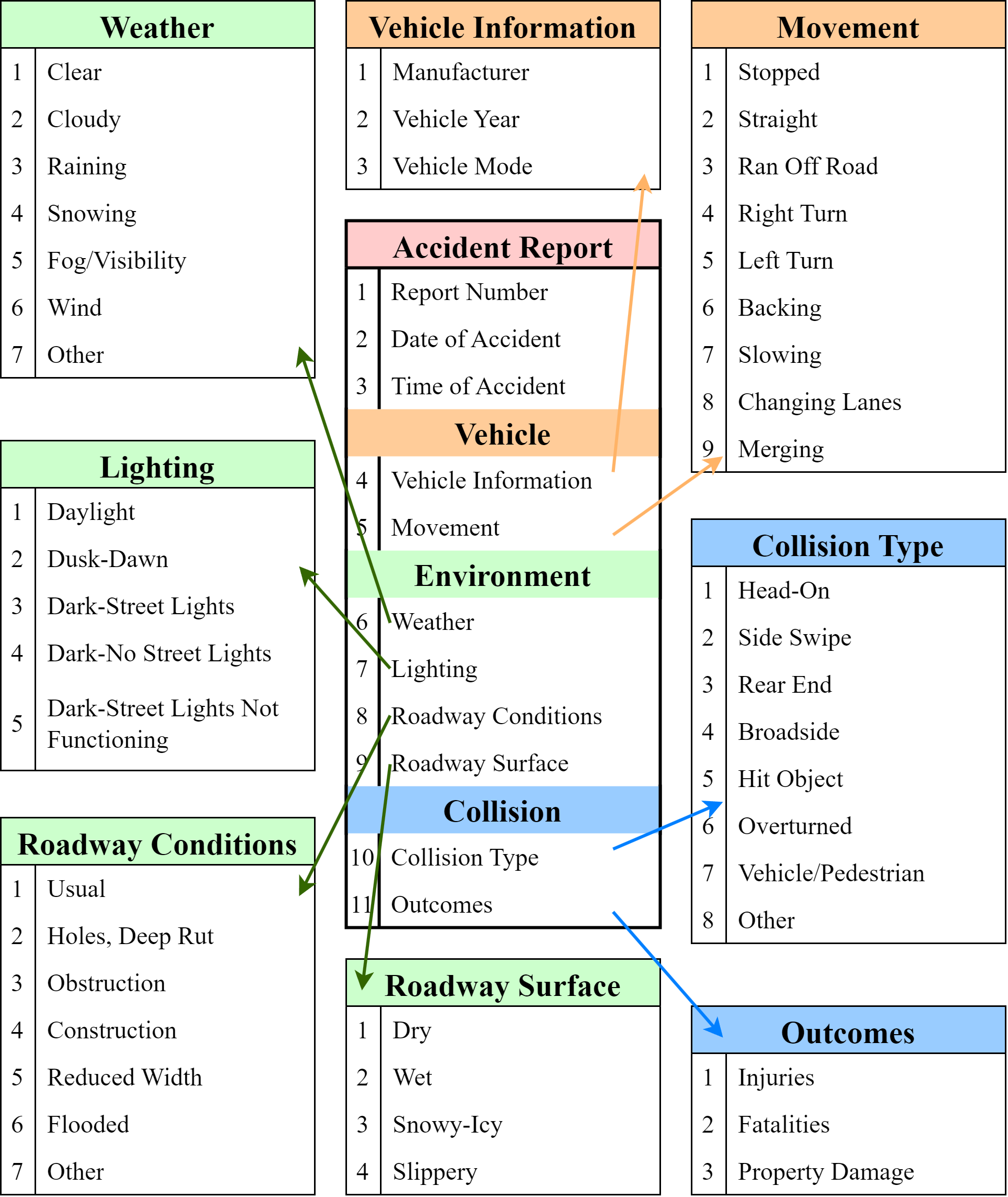}
 
\caption{ { {Contributing factors in traffic accident reports. The key contributing factors can be divided into three parts: before the accident, during the accident, and after the accident. The main contributing factors before the accident are environmental information, including weather, lighting, road conditions, etc. The main contributing factors during the accident are concentrated on the key information of the accident participants, such as vehicle information, vehicle behavior, etc. The contributing factors after the accident are mainly the results of the accident, such as the type of accident, the severity of the accident, etc.}}}
\label{Fig.4}
\end{figure}

\textbf{(1) Accident Reports.} We collected 573 autonomous vehicle accident reports from the California Department of Motor Vehicles (DMV), spanning from 2019 to mid-2024, to serve as positive samples for our model training. These reports are part of a regulatory requirement in California DMV, where manufacturers testing autonomous vehicles on public roads must submit detailed accounts of any incidents involving their vehicles. As shown in Fig. \ref{Fig.3} and Fig. \ref{Fig.4}, each report includes comprehensive data on various factors, such as weather conditions, lighting, roadway conditions, vehicle information, collision types, and outcomes, including any injuries or fatalities. This wealth of information allows us to meticulously reconstruct each accident scenario, capturing the complex interactions between environmental conditions, vehicle behavior, and the resulting collisions. These reconstructed scenarios are then encoded using a VLM model, providing a nuanced dataset that enhances the accuracy and depth of our traffic accident anticipation model.

\textbf{(2) Non-Accident Reports.} Given the absence of detailed textual reports for non-accident scenarios, we generated 500 negative samples using GPT-4o \cite{achiam2023gpt}. The creation of these reports followed a carefully designed and automated process, ensuring consistency and relevance to real-world driving conditions.

\textit{Unified Prompt Design.}
To generate a large dataset of non-crash reports efficiently, a single unified prompt was crafted to encompass a wide range of safe driving scenarios. This prompt was designed to simulate normal driving behavior under various conditions, reflecting the same categories as accident reports (e.g. weather, lighting, road conditions) but with an emphasis on safety.

\textsf{``Based on the format and content of the input reports and the list of input traffic contributing factors, generate 500 traffic scenario reports where vehicles navigate \textbf{[varied road conditions]} and \textbf{[different environmental factors]} from the list while adhering to all safety protocols. Ensure the scenarios reflect a wide range of normal driving behaviors, avoiding critical errors or hazardous actions. Each report should describe the scenario in detail, including vehicle movements, environmental conditions, and the absence of accidents.''}

This prompt (short version) allows for the automatic generation of 500 unique reports that cover a broad spectrum of safe driving situations, ensuring that the non-crash scenarios are diverse and realistic.

\textit{Automated Generation and Evaluation.}
The generation of non-crash reports was fully automated. After generating the reports, a secondary evaluation process was employed to ensure the quality and relevance of the output. This process involved the following steps:

\begin{itemize}
    \item \textbf{Post-Generation Filtering}: After the initial batch of 500 reports was generated, each report was automatically evaluated with GPT-4o and Claude 3.5 to ensure it met the criteria for non-crash scenarios. Reports that contained any descriptions of risky behavior or potential accident indicators were flagged and reprocessed to align with the intended non-accident focus.
    \item \textbf{Consistency Check}: An automated consistency check was implemented to ensure that each report adhered to the safety emphasis described in the unified prompt. This step involved comparing generated reports against the accident scenarios to confirm the absence of any factors typically associated with crashes.
\end{itemize}

\textit{Human Expert Review.} Finally, the generated reports need to be reviewed by human experts and their validity confirmed. The human expert review focuses on the following aspects:
The generated reports must conform to the real logic, and there must not be cases where the environment and driving behavior conflict, such as high-speed driving in foggy or rainy days.
Besides, since LLM generates text word by word through chained conditional probabilities, local optimal traps may occur. Human experts need to check whether there is a large amount of duplication in the generated content to ensure the independence of different reports.

Through context constraints, domain knowledge constraints, consistency checks and manual reviews, the generated traffic reports can avoid the impact of potential hallucinations of LLM to the greatest extent, eliminate dangerous factors and be consistent with the logic of real life.

\textit{Key Differences.}
The generated non-crash reports are distinguished from crash reports by their emphasis on safe and controlled driving behaviors. For example, while a crash report might describe a vehicle failing to yield in poor weather conditions leading to a collision, a corresponding non-crash report would depict the vehicle successfully navigating the same conditions by adhering to safe driving practices. This distinction is crucial for training the model to recognize both risky and safe behaviors.

\begin{itemize}
    \item \textbf{Crash Report Example (Part)}: ``The vehicle failed to slow down in wet conditions, leading to a rear-end collision.''
    \item \textbf{Non-Crash Report Example (Part)}: ``The vehicle reduced speed and maintained a safe following distance while navigating the wet road, avoiding potential incidents.''
\end{itemize}

Both crash and non-crash reports are encoded using the same vision-language model, enabling the model to distinguish critical differences between scenarios that result in accidents and those that do not. This dual-dataset approach enhances the model's predictive capability by identifying both risk factors and behaviors that mitigate accidents.

The systematic generation and evaluation of non-crash reports ensure consistency and alignment with the objectives of accurate traffic accident anticipation. By automating this process, we maintain rigorous quality control across the dataset, providing a robust foundation for model training.

\begin{table*}[]
\centering
\footnotesize
\caption{{Distribution of accident damage across various contributing factors in the collected dataset. The dataset encompasses a diverse range of traffic scenarios, revealing significant correlations between contributing factors and accident severity. For example, accidents occurring under dusk or nighttime conditions tend to be more severe than those in daylight, and broadside collisions result in greater injuries compared to Head-On collisions. These insights enable the identification of high-risk factors and the prioritization of predicted accidents based on their urgency and anticipated impact.}}
   \resizebox{\linewidth}{!}{
\begin{tabular}{ccccccccccc}
\hline
\multicolumn{2}{c}{Accident Related Factors} &
  \multicolumn{2}{c}{None} &
  \multicolumn{2}{c}{Minor} &
  \multicolumn{2}{c}{Moderate} &
  \multicolumn{2}{c}{Major} &
  Total \\ \hline
\multirow{10}{*}{Damage Area}      & Front Bumper               & 1  & 2.27\%   & 27  & 61.36\%  & 14 & 31.82\%  & 2  & 4.55\%  & 44  \\
                                   & Front Driver Side          & 0  & 0.00\%   & 36  & 52.94\%  & 27 & 39.71\%  & 5  & 7.35\%  & 68  \\
                                   & Front Passenger Side       & 0  & 0.00\%   & 40  & 76.92\%  & 9  & 17.31\%  & 3  & 5.77\%  & 52  \\
                                   & Left Front Corner          & 0  & 0.00\%   & 72  & 60.50\%  & 45 & 37.82\%  & 2  & 1.68\%  & 119 \\
                                   & Left Rear                  & 1  & 0.55\%   & 135 & 73.77\%  & 39 & 21.31\%  & 8  & 4.37\%  & 183 \\
                                   & Left Rear Passenger        & 0  & 0.00\%   & 23  & 51.11\%  & 18 & 40.00\%  & 4  & 8.89\%  & 45  \\
                                   & Rear Bumper                & 4  & 2.31\%   & 131 & 75.72\%  & 30 & 17.34\%  & 8  & 4.62\%  & 173 \\
                                   & Right Front Corner         & 1  & 0.92\%   & 67  & 61.47\%  & 36 & 33.03\%  & 5  & 4.59\%  & 109 \\
                                   & Right Rear                 & 0  & 0.00\%   & 130 & 76.02\%  & 32 & 18.71\%  & 9  & 5.26\%  & 171 \\
                                   & Right Rear Passenger       & 0  & 0.00\%   & 26  & 65.00\%  & 10 & 25.00\%  & 4  & 10.00\% & 40  \\
\hline                                   
\multirow{4}{*}{Weather}           & Clear                      & 46 & 9.33\%   & 364 & 73.83\%  & 73 & 14.81\%  & 10 & 2.03\%  & 493 \\
                                   & Cloudy                     & 5  & 8.20\%   & 43  & 70.49\%  & 10 & 16.39\%  & 3  & 4.92\%  & 61  \\
                                   & Raining                    & 1  & 3.57\%   & 21  & 75.00\%  & 5  & 17.86\%  & 1  & 3.57\%  & 28  \\
                                   & Fog/Visibility             & 0  & 0.00\%   & 2   & 100.00\% & 0  & 0.00\%   & 0  & 0.00\%  & 2   \\
                                 \hline           
\multirow{2}{*}{Roadway}           & Dry                        & 44 & 8.46\%   & 384 & 73.85\%  & 80 & 15.38\%  & 12 & 2.31\%  & 520 \\
                                   & Wet                        & 4  & 10.26\%  & 28  & 71.79\%  & 5  & 12.82\%  & 2  & 5.13\%  & 39  \\
                                 \hline           
\multirow{7}{*}{Road Conditions}   & Holes                      & 0  & 0.00\%   & 2   & 100.00\% & 0  & 0.00\%   & 0  & 0.00\%  & 2   \\
                                   & Loose Material on Roadway  & 0  & 0.00\%   & 1   & 50.00\%  & 1  & 50.00\%  & 0  & 0.00\%  & 2   \\
                                   & Obstruction                & 0  & 0.00\%   & 3   & 60.00\%  & 2  & 40.00\%  & 0  & 0.00\%  & 5   \\
                                   & Construction               & 1  & 20.00\%  & 4   & 80.00\%  & 0  & 0.00\%   & 0  & 0.00\%  & 5   \\
                                   & Reduced Roadway Width      & 0  & 0.00\%   & 4   & 100.00\% & 0  & 0.00\%   & 0  & 0.00\%  & 4   \\
                                   & Other Unusual              & 0  & 0.00\%   & 6   & 66.67\%  & 2  & 22.22\%  & 1  & 11.11\% & 9   \\
                                   & Usual                      & 48 & 9.18\%   & 383 & 73.23\%  & 79 & 15.11\%  & 13 & 2.49\%  & 523 \\
                                   \hline           
\multirow{4}{*}{Lighting}          & Daylight                   & 41 & 10.41\%  & 299 & 75.89\%  & 50 & 12.69\%  & 4  & 1.02\%  & 394 \\
                                   & Dusk-Dawn                  & 2  & 12.50\%  & 12  & 75.00\%  & 1  & 6.25\%   & 1  & 6.25\%  & 16  \\
                                   & Dark-Street Lights         & 9  & 5.39\%   & 115 & 68.86\%  & 34 & 20.36\%  & 9  & 5.39\%  & 167 \\
                                   & Dark-No Street Lights      & 0  & 0.00\%   & 1   & 25.00\%  & 3  & 75.00\%  & 0  & 0.00\%  & 4   \\
                                   \hline           
\multirow{15}{*}{Movement}         & Stopped                    & 20 & 8.55\%   & 180 & 76.92\%  & 28 & 11.97\%  & 6  & 2.56\%  & 234 \\
                                   & Proceeding Straight        & 13 & 7.43\%   & 120 & 68.57\%  & 37 & 21.14\%  & 5  & 2.86\%  & 175 \\
                                   & Making Right Turn          & 4  & 11.43\%  & 27  & 77.14\%  & 4  & 11.43\%  & 0  & 0.00\%  & 35  \\
                                   & Making Left Turn           & 2  & 5.71\%   & 25  & 71.43\%  & 8  & 22.86\%  & 0  & 0.00\%  & 35  \\
                                   & Making U Turn              & 0  & 0.00\%   & 1   & 100.00\% & 0  & 0.00\%   & 0  & 0.00\%  & 1   \\
                                   & Backing                    & 5  & 20.00\%  & 18  & 72.00\%  & 1  & 4.00\%   & 1  & 4.00\%  & 25  \\
                                   & Slowing/Stopping           & 6  & 12.00\%  & 36  & 72.00\%  & 6  & 12.00\%  & 2  & 4.00\%  & 50  \\
                                   & Passing Other Vehicle      & 0  & 0.00\%   & 1   & 100.00\% & 0  & 0.00\%   & 0  & 0.00\%  & 1   \\
                                   & Changing Lanes             & 0  & 0.00\%   & 9   & 69.23\%  & 4  & 30.77\%  & 0  & 0.00\%  & 13  \\
                                   & Parking Manuever           & 1  & 10.00\%  & 9   & 90.00\%  & 0  & 0.00\%   & 0  & 0.00\%  & 10  \\
                                   & Entering Traffic           & 1  & 20.00\%  & 4   & 80.00\%  & 0  & 0.00\%   & 0  & 0.00\%  & 5   \\
                                   & Driving into Opposing Lane & 0  & 0.00\%   & 1   & 100.00\% & 0  & 0.00\%   & 0  & 0.00\%  & 1   \\
                                   & Parked                     & 2  & 15.38\%  & 9   & 69.23\%  & 2  & 15.38\%  & 0  & 0.00\%  & 13  \\
                                   & Merging                    & 0  & 0.00\%   & 3   & 75.00\%  & 1  & 25.00\%  & 0  & 0.00\%  & 4   \\
                                   & Other                      & 0  & 0.00\%   & 4   & 80.00\%  & 1  & 20.00\%  & 0  & 0.00\%  & 5   \\
                                   \hline           
\multirow{7}{*}{Type of collision} & Head-On                    & 4  & 10.53\%  & 25  & 65.79\%  & 8  & 21.05\%  & 1  & 2.63\%  & 38  \\
                                   & Side Swipe                 & 5  & 6.10\%   & 62  & 75.61\%  & 14 & 17.07\%  & 1  & 1.22\%  & 82  \\
                                   & Rear End                   & 15 & 13.89\%  & 78  & 72.22\%  & 12 & 11.11\%  & 3  & 2.78\%  & 108 \\
                                   & Broadside                  & 2  & 9.09\%   & 12  & 54.55\%  & 5  & 22.73\%  & 3  & 13.64\% & 22  \\
                                   & Hit Object                 & 0  & 0.00\%   & 30  & 76.92\%  & 7  & 17.95\%  & 2  & 5.13\%  & 39  \\
                                   & Vehicle/Pedestrian         & 1  & 100.00\% & 0   & 0.00\%   & 0  & 0.00\%   & 0  & 0.00\%  & 1   \\
                                   & Other                      & 4  & 16.00\%  & 19  & 76.00\%  & 2  & 8.00\%   & 0  & 0.00\%  & 25  \\
                                   \hline           
\multirow{10}{*}{Other Factors} &
  CVC Sections Violated &
  1 &
  33.33\% &
  1 &
  33.33\% &
  1 &
  33.33\% &
  0 &
  0.00\% &
  3 \\
                                   & Vision Obscurement         & 0  & 0.00\%   & 1   & 50.00\%  & 1  & 50.00\%  & 0  & 0.00\%  & 2   \\
                                   & Inattention                & 2  & 20.00\%  & 6   & 60.00\%  & 2  & 20.00\%  & 0  & 0.00\%  & 10  \\
                                   & Stop \& Go Traffic         & 1  & 9.09\%   & 8   & 72.73\%  & 1  & 9.09\%   & 1  & 9.09\%  & 11  \\
                                   & Entering/Leaving Ramp      & 0  & 0.00\%   & 6   & 66.67\%  & 3  & 33.33\%  & 0  & 0.00\%  & 9   \\
                                   & Unfamiliar With Road       & 0  & 0.00\%   & 0   & 0.00\%   & 1  & 100.00\% & 0  & 0.00\%  & 1   \\
                                   & Uninvolved Vehicle         & 0  & 0.00\%   & 1   & 100.00\% & 0  & 0.00\%   & 0  & 0.00\%  & 1   \\
                                   & None Apparent              & 1  & 4.35\%   & 20  & 86.96\%  & 2  & 8.70\%   & 0  & 0.00\%  & 23  \\
                                   & Runaway Vehicle            & 1  & 50.00\%  & 0   & 0.00\%   & 1  & 50.00\%  & 0  & 0.00\%  & 2   \\
                                   & Other                      & 0  & 0.00\%   & 1   & 100.00\% & 0  & 0.00\%   & 0  & 0.00\%  & 1  \\ \hline
\end{tabular}
}
\label{tab.1}
\end{table*}

\textbf{(3) Text Encoder.} 
The encoder leverages Long-CLIP \cite{zhang2024long}, a model capable of processing lengthy texts, to encode all accident reports. It employs max-pooling techniques to compress the long-text data, ensuring the preservation of significant semantic information within the text. This process ultimately generates the textual features \(X = \{X_{pos}, X_{neg}\}\), where \(X_{pos}\) represents the accident-related features and \(X_{neg}\) corresponds to the non-accident features.
\begin{equation}
X=\phi_{\textit{max-pooling}}(\phi_{\textit{nm}}(\phi_{\textit{Long-CLIP}}({Y})) )
\end{equation}
where $\phi_{\textit{Long-CLIP}}$ is Long-CLIP model,  $\phi_{\textit{nm}}$ is the normalization, and \({Y}\) represents all related reports.

Accident-related features are derived from real traffic accident reports, describing dangerous situations that may cause traffic accidents; non-accident features are derived from artificially generated non-accident scenario texts, describing typical scenarios of safe vehicle driving. These text features are encoded into high-dimensional vectors through Long-CLIP to align and compare with visual features extracted from videos, effectively combining historical experience and real-time visual information to measure the similarity between the current video scene and historical accident or non-accident situations.

\subsection{Step-3: Knowledge Injection and Accident Anticipation}

In this step, we propose a novel loss function that supervises the model in learning domain knowledge while utilizing the frame features and textual features obtained from the first and second steps for accident detection. Specifically, we first apply a multi-head attention mechanism to capture frame-wise pyramid feature maps from the frame features $V_f$, resulting in vision features \( O_f \in \mathbb{R}^{ N \times D} \):

\begin{equation}
\begin{cases}
{Q}_v={W}^{Q}_{v} (\phi_{\textit{nm}}(V_f)), \\
{K}_{v}={W}^{K}_{v} (\phi_{\textit{nm}}(V_f))\\
{V}_{v}={W}^{V}_{v}  (\phi_{\textit{nm}}(V_f))
\end{cases}
\end{equation}
such that,
\begin{equation}
O_f={\phi_{\text{Softmax}}}\left(\frac{{Q}_t ({K}_v)^{\top}}{\sqrt{d_k}}\right) V_v 
\end{equation}
where $\phi_{\text{Softmax}}$ is the Softmax activation function. Next, these features are fed into a linear layer to generate confidence logits \( O \in \mathbb{R}^{N \times 2} \), which indicate the probability that the vision features belong to either the non-accident or accident categories, expressed as $O=\phi_{\textit{MLP}}(O_f)$. Our model will classify a video as positive if $O_{n,1}$ is greater than a learnable threshold.

To capture the relationship between the sequence of visual features and classification logits, we use visual contexts to refine the class embeddings, as these contexts can enhance the accuracy of succinct text labels. To achieve this, we introduce an anomaly-focused visual prompt, which concentrates on visual embeddings within abnormal segments and aggregates these embeddings as video-level prompts for class embeddings \cite{wu2024vadclip}.
\begin{equation}
V_{\text{attn}} = \phi_{\textit{nm}} \left(O^T \cdot V\right)
\end{equation}

Here, \(V_{\text{attn}} \in \mathbb{R}^{B \times C \times D}\), \(O \in \mathbb{R}^{B \times N \times C}\) (output logits) and \(V \in \mathbb{R}^{B \times N \times D}\) (visual features), \(B\) represents the batch size, and \(T\) represents the transpose operation.
Subsequently, we integrate \(V_{\text{attn}}\) with the class embedding \(X\), resulting in the final instance-specific class embedding, denoted as \(I = \textit{FFN} (\textit{ADD}(V_{\text{attn}}, X)) + X\).

Finally, the dot product of the normalized visual and textual features is computed, and scaled by a temperature parameter, which is set to 0.07, to obtain similarity scores \(S\) between the visual and textual features.
\begin{equation}
S = \frac{X_{\text{norm}} \cdot \frac{I + \textit{MLP}(I)}{\|I + \textit{MLP}(I)\|_{2}}
^T}{0.07}
\end{equation}
With the accident confidence $O$ and similarity score $S$, we proposed that our loss function would comprise three components. The classification loss \(L_{ce}\) is a video-level loss obtained by performing max pooling over all frames preceding an accident in the video to generate a video-level prediction, followed by calculating the cross-entropy loss between this prediction and the target label.
\begin{equation}
L_{ce} = \sum_{i=1}^{B} \phi_{\textit{CE}}\left(\max({O}_{i, 1:toa, :}\mathbf{L}_{i, 1}\right)
\end{equation}
where \(O \in \mathbb{R}^{B \times F \times 2}\) is the prediction output logits, \(B\) is the batch size, \(\mathbf{L} \in \mathbb{R}^{B \times 2}\) is the one-hot encoded ground truth labels, \(\phi_{\textit{CE}}\) is cross-entropy loss. Furthermore, the prediction loss \(L_t\) is a frame-level custom loss based on an exponential penalty of the time difference to the accident. For the current time step \(t\) and the accident occurrence time \(\tau\), the time-based penalty can be calculated as follows:
\begin{equation}
\text{penalty} = -\max\left(0, \frac{\tau - t - 1}{\text{fps}}\right)
\end{equation}
\begin{equation}
L_{\text{pos}} = -\exp(\text{penalty}) \cdot \phi_{\textit{CE}}(O, \mathbf{L}_{i, 1})
\end{equation}
\begin{equation}
L_{\text{neg}} = \phi_{\textit{CE}}(O, \mathbf{L}_{i, 1})
\end{equation}  
Then, the total prediction loss \(L_t\) is formulated as follows:
\begin{equation}
    L_t = \frac{1}{N} \sum_{i=1}^{B} \sum_{j=1}^{N} \left[ L_i^{(1)} \cdot L_{\text{pos}} + L_i^{(0)} \cdot L_{\text{neg}} \right]
\end{equation} 

The multi-instance learning loss \(L_{mil}\) assesses the accuracy of similarity-based predictions by classifying the most representative frames within each video and calculating the corresponding classification loss.
For each video \(i\), select the top \(k\)=20 frame logits and calculate their mean:
\begin{equation}
    \mathbf{I}_i = \frac{1}{k} \sum_{j=1}^{k} \text{topk}(S_i, k)
\end{equation}
where \(L_{mil}\) is computed by taking the sum of the product of the normalized labels and the log-softmax of the instance logits, followed by averaging this sum across all instances.
\begin{equation}
    L_{mil} = -\frac{1}{B} \sum_{i=1}^{B} \sum_{j=1}^{C} L_{ij} \log \left( \frac{\exp(\mathbf{I}_{ij})}{\sum_{c=1}^{C} \exp(\mathbf{I}_{ic})} \right)
\end{equation}
where \(C\)=2 represents positive and negative classes.

\subsection{Step-4: Accident Alert Feedback}
This step primarily involves utilizing the LLM to generate accident warnings and driving recommendations. We propose a novel approach to fine-tune the LLM by jointly incorporating images, model prediction results, and accident-related factors. Specifically, in this study, to ensure that GPT-4o accurately interprets the input traffic scenes and effectively conveys the model's output, we guided the LLM's attention by incorporating the model's prediction results alongside contributing factors in traffic accident reports when inputting the scene images. 

When designing prompts for this step with the LLM, we incorporated the following critical considerations:

 {\textbf{Stage 1: Risk Assessment.}}  We developed a risk assessment model that integrates accident probability scores and raw model inputs to estimate the urgency of each collision scenario. The analysis involves evaluating the type and number of agents present in the environment (e.g., vehicles, trucks, pedestrians, cyclists) and their relative distances to the self, combined with the accident probability scores to determine urgency.

{\textbf{Stage 2: Human Compatibility.}} The alerts generated by the model are dynamically tailored in terms of tone, length, and content based on the assessed risk urgency level. This adaptive approach is designed to enhance the passenger experience and provide greater transparency in algorithmic decision-making. For instance, during normal driving conditions, the model delivers friendly reminders, whereas in high-risk situations, it offers urgent and direct feedback to ensure timely response.

{\textbf{Stage 3: Legality Verification.}} To ensure compliance with traffic laws and regulations, all alerts generated by the LLM undergo a rigorous legality verification process. This process systematically cross-references each alert against a comprehensive database of traffic laws and legal standards, ensuring that the model’s recommendations adhere to regulatory requirements and do not result in legal violations.

Our prompt design guides the LLM in analyzing the overall traffic scenario and identifying key contributing factors within the scene. Based on the accident anticipation results and the urgency of the situation, the LLM outputs alerts classifies accident types, provides driving recommendations, and conducts scene analysis in a prioritized manner. 

Here, we present LLM feedback in its entirety corresponding to the video scene in Fig. \ref{Fig.1}:

``Potential accident detected. Be prepared to slow down, maintain a safe distance, and watch out for sudden movements from the motorcycle on the right. The road appears to be clear and dry, with normal urban traffic.
A silver car on the left. A black car ahead might be turning now, and a motorcycle is moving forward close to the curb, which could lead to a side collision if it swerves or keeps going.''

\begin{figure}[t]
\centering  
\includegraphics[width=0.45\textwidth]{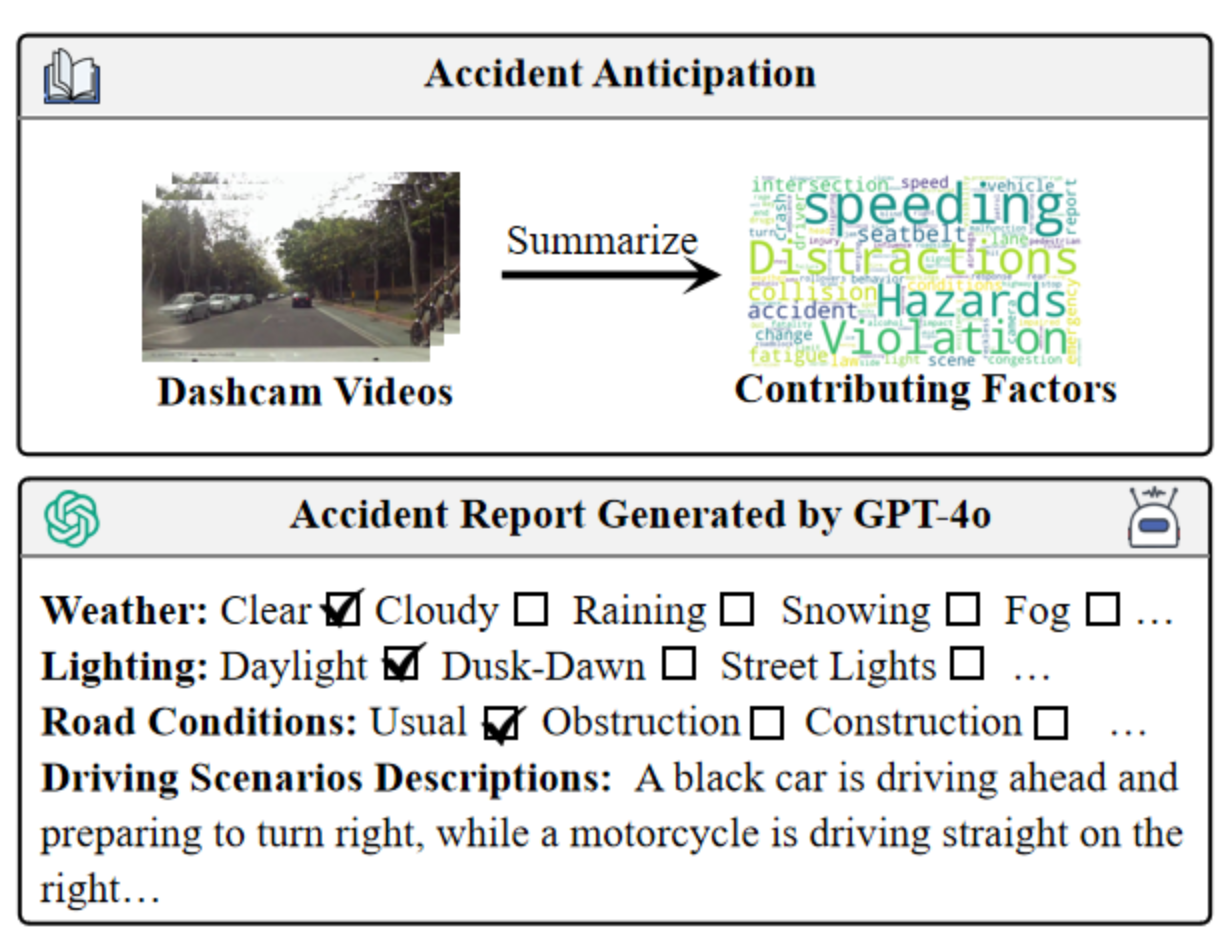}
\caption{{Example of accident archiving for the accident scene in Fig. \ref{Fig.1}. Leveraging information extracted from the accident anticipation model, a large language model generates standardized accident reports, ensuring a structured and comprehensive documentation of the incident.}}
\label{Fig.5}
\end{figure}

\subsection{Post-step: Accident Archiving}
During the accident anticipation process, key accident factors from the current accident scene are extracted. To leverage this information effectively, we propose a unified framework for accident report utilization and archiving, as illustrated in Fig. \ref{Fig.5}. This framework utilizes existing accident reports to enhance accident anticipation while simultaneously generating new reports for predicted accidents.

Before an accident occurs, the model continuously monitors and records environmental factors surrounding the ego vehicle, including weather conditions, lighting, road surface status, and location. This information is temporarily stored in short-term memory for real-time accident anticipation. When an accident is predicted or detected, the system logs critical data, such as accident participants and vehicle behaviors.
Based on the extracted key accident factors, an LLM generates a structured accident description and compiles a comprehensive accident report, which is then archived.

This framework extends the functionality of our accident anticipation model, significantly reducing the manpower required for accident documentation, and has great deployment potential in key transportation system agencies such as real-world intelligent transportation systems, autonomous driving platforms, smart city infrastructure, traffic management centers, and insurance claims processing centers. In addition, by continuously archiving predicted or detected accidents, the system can build a growing dataset for incremental training. Therefore, the accident anticipation capability will improve over time, promoting a positive cycle of data collection and application, and improving overall road safety and efficiency.

Despite the many benefits, the proposed framework also poses several risks. First, the collection and archiving of detailed accident data raises concerns about data privacy. Therefore, strict anonymization and encryption techniques must be implemented to protect sensitive personal information. Second, false positives generated by the model may lead to incorrect accident reports being archived, which may reduce the quality of subsequent training data. To mitigate this, verification and validation with official corresponding accident reports is required, which may involve manual verification or additional automated checks. Finally, accident data archiving must strictly comply with local and international data privacy regulations (such as GDPR) to ensure ethical and legal operations \cite{gdpr2018general,collingwood2017privacy}.

\subsection{Learnable Threshold}

Traditional systems employ fixed threshold (e.g., $\tau=0.5$) or grid-search optimized threshold, which suffer from some fundamental limitations:
\begin{itemize}
\item Fixed thresholds are static and cannot automatically adapt to the dynamic feature distribution of different samples and time steps. When the ratio of positive and negative samples is very different (such as the ratio of positive samples to negative samples in the DAD dataset is 1 to 5), fixed thresholds are prone to lead to excessively high false positive rates.

\item The grid-search optimized threshold is limited by the discretization step size, and the optimal threshold is calculated by exhaustive method, which is decoupled from the model training process.
\end{itemize}

To this end, we propose a learnable threshold mechanism, which integrates the threshold parameters into the model training process to achieve dynamic and adaptive decision boundary optimization.

Let $z_t = [z_t^{(0)}, z_t^{(1)}] \in \mathbb{R}^2$ denote the raw logits at timestep $t$, where $z_t^{(1)}$ represents the positive class confidence. We introduce a learnable threshold parameter $\tau \in \mathbb{R}$ with the following adjustment:

\begin{equation}
    \tilde{z}_t^{(1)} = z_t^{(1)} - \tau
    \label{eq:threshold_adjust}
\end{equation}
$\tau$ is initialized to 0, at which time $\tilde{z}_t^{(1)} = z_t^{(1)}$, which is equivalent to the traditional method $\tau_{fixed}=0.5$.
The classification decision rule becomes:
\begin{equation}
    \text{Positive} \iff \text{softmax}(\tilde{\mathbf{z}}_t)_1 > 0.5 \iff \tilde{z}_t^{(1)} > z_t^{(0)}
    \label{eq:decision_rule}
\end{equation}

The gradient flow through the threshold parameter follows:
\begin{align}
    \frac{\partial \mathcal{L}}{\partial \tau}= -\sum_{t=1}^T \frac{\partial \mathcal{L}}{\partial \tilde{z}_t^{(1)}}= -\frac{1}{N}\sum_{t=1}^T (y_t - p_t)
    \label{eq:gradient_flow}
\end{align}
where $p_t = \sigma(z_t^{(1)} - z_t^{(0)} - \tau)$ is the adjusted prediction probability, $\sigma$ represents the Sigmoid function and $\mathcal{L}$ donates the total loss function. This reveals the self-calibration property: $\tau$ increases when over-confident ($p_t > y_t$) and decreases when under-confident ($p_t < y_t$).

\begin{figure}[h]
\centering  
\includegraphics[width=0.4\textwidth]{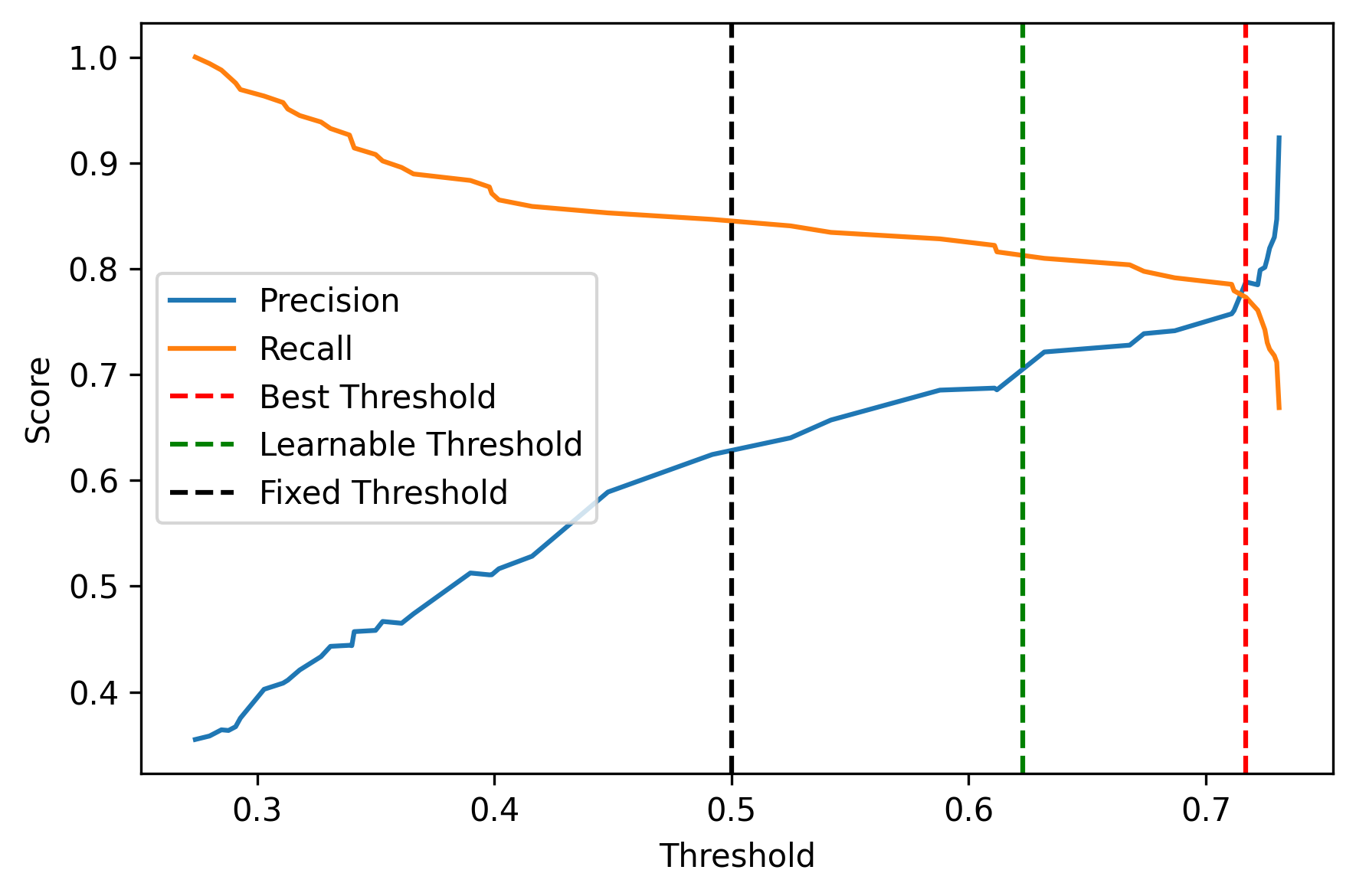}

\caption{{Precision and recall curve and threshold comparison on DAD testing dataset. Best Threshold represents the threshold with the highest F1 score selected by exhaustive method during the testing process.}}
\label{Fig.6}
\end{figure}

Fig. \ref{Fig.6} shows the curves of the accuracy and recall of the same model during the testing process as the threshold changes. Best threshold is the threshold with the highest F1 score, that is, the threshold position that best balances accuracy and recall, while the fixed threshold represents the threshold with 0.5 as the demarcation used in traditional research. It can be seen that our learnable threshold is closer to the balance state of accuracy and recall than the fixed threshold, and can effectively reduce false positives in the current testing process.

\section{Experiments}
\label{Experiments}

\begin{table*}[htbp]
    \centering
\caption{{Comparison of the DAD, CCD, and A3D datasets. In contrast to A3D, the data distribution in DAD and CCD is more balanced. Additionally, DAD features a higher frame rate and a longer pre-accident time range, making it the primary dataset for this study.}}

    \setlength\tabcolsep{6pt}
     \resizebox{0.8\linewidth}{!}{
    \begin{tabular}{cccccccc}
    \hline
        Dataset & Videos & Positives & Frames& FPS &Accident Time  &Ego-Involved &Accident Reasons\\ \hline
        DAD & 1750 & 620  &100& 20 & Fixed &\checkmark &~\\ 
        CCD & 4500 & 1500  &50& 10 & Random &\checkmark &~\\ 
        A3D & 1500 & 1500  &100& 20 & Random  &\checkmark &\checkmark\\ \hline
    \end{tabular}
    }

    \label{datasets}
    
\end{table*}

\subsection{Datasets}
As shown in Table \ref{datasets}, this study evaluates the performance of our model on three real-world datasets:

\textbf{DAD} The Dashcam Accidents Dataset (DAD) \cite{chan2017anticipating} consists of 1,750 dashcam videos collected from six cities in Taiwan. It includes 620 accident (positive) and 1,130 non-accident (negative) videos. The dataset is split into 1,284 training samples (455 positives, 829 negatives) and 466 testing samples (165 positives, 301 negatives). Each video spans 5 seconds (100 frames), with accident occurrences aligned to the 90th frame in accident sequences.

\textbf{CCD} The Car Crash Dataset (CCD) \cite{BaoMM2020} comprises 4,500 videos, including 1,500 accident videos annotated from YouTube and 3,000 non-accident videos sourced from the BDD100K dataset \cite{yu2020bdd100k}. Each video lasts 5 seconds (50 frames), with accident events randomly positioned within the final two seconds for accident videos. The dataset also provides coarse annotations on accident causes, weather conditions, and environmental factors. CCD is partitioned into 3,600 training samples (1,200 positives, 2,400 negatives) and 900 testing samples (300 positives, 600 negatives).

\textbf{A3D} The An An Accident Detection (A3D) dataset \cite{yao2019unsupervised} contains 1,500 dashcam video clips from diverse East Asian urban environments, covering various weather conditions and times of day. We utilize the pre-extracted video features provided by Bao et al. \cite{BaoMM2020}, captured at 20 FPS, with each clip containing 100 frames. The dataset is divided into training and testing sets in an 80\%-20\% ratio. Originally designed for Traffic Anomaly Detection, A3D contains only accident videos. Negative samples are generated from non-accident segments, allowing its application in traffic accident anticipation tasks.

\subsection{Evaluation Metrics}

To assess the timeliness and accuracy of the model’s anticipation,  Average Precision (AP) and the Mean Time-To-Accident (mTTA) are utilized in this study. 

\textbf{Average Precision.} This is a metric used to evaluate the accuracy of predicting traffic events from dashcam videos. AP measures the model's ability to correctly identify the occurrence of traffic accidents, particularly in cases with an imbalance between positive and negative samples. This is done by assessing both precision and recall across various threshold settings.
In binary classification tasks, precision (P) is the ratio of true positives (TP) to the sum of true positives (TP) and false positives (FP), while recall (R) is the ratio of true positives (TP) to the sum of true positives (TP) and false negatives (FN). The precision-recall curve is plotted using these values, and AP is defined as the area under this curve. For a given threshold $p$, if the confidence score $a_{t}^{p}$ at time $t$ exceeds this threshold, a traffic event is predicted. The resulting classifications are categorized as TP, FP, FN, and TN. The AP is then calculated by integrating the precision-recall curve, which can be approximated by discrete summation. Formally,
\begin{equation}
AP = \int P(R) \, dR = \sum_{k=0}^{m} P(k) \Delta R(k)
\end{equation}

\textbf{Mean Time-To-Accident.} It quantifies a model's ability to predict the occurrence of an accident among positive samples in advance. This metric evaluates the earliness of accident anticipation based on positive predictions from dashcam videos. If an accident is predicted to happen at the $x$-th frame and occurs at the $y$-th frame, TTA is defined as $y$-$x$.  By analyzing different threshold values $p$, sequences of TTA results and corresponding recall rates can be derived. The average of these TTA values is termed mTTA. Nonetheless, it is crucial to recognize that a high TTA might be misleading if the model overfits the dashcam data, resulting in indiscriminate positive predictions. Hence, a high TTA value is ineffective without considering the AP. This study reports the TTA values when the highest AP is attained, ensuring a comprehensive evaluation of the model's predictive performance.

\subsection{Experiment Setups}
Our model is implemented in PyTorch and trained on an NVIDIA GeForce RTX 4090 GPU. We use the ReduceLROnPlateau scheduler for adaptive learning rate adjustment, with a reduction factor of 0.5 and a patience threshold of 3 epochs. The model is trained for 30 epochs with a batch size of 10. The Adam optimizer is employed with an initial learning rate of \(1 \times 10^{-3}\) (the initial learning rate of learnable threshold is \(1 \times 10^{-1}\)),  and L2 regularization is applied to mitigate overfitting, using a weight decay of \(1 \times 10^{-4}\). The loss function is initialized with equal weight ratios (1:1:1), and these weights are treated as a learnable tensor.

Notably, in the feature extraction stage, we set the embedding dimension for VGG-16 to 512. For Long-CLIP, we use the pre-trained longclip-L model with a default embedding dimension of 768, which is subsequently reduced to 512 via a linear transformation to maintain consistency with the visual feature embeddings.

\subsection{Evaluation Results}
\begin{table}[t]
\centering
\caption{ {Comparison of models seeking \textbf{balance} between mTTA and AP on DAD, CCD and A3D. \textbf{Bold} and \underline{underlined} values represent the best and second-best performance. ``$\uparrow$'' indicates that higher values correspond to improved model performance in each category. Instances where values are not available are marked with a dash (``-'').}}
 \resizebox{\linewidth}{!}{
\begin{tabular}{ccccccc}
\hline \specialrule{0em}{1pt}{1pt}
\multirow{2}[2]{*}{Model} & \multicolumn{2}{c}{DAD} & \multicolumn{2}{c}{CCD} & \multicolumn{2}{c}{A3D} \\
\cmidrule(lr){2-3} \cmidrule(lr){4-5} \cmidrule(lr){6-7}
 & AP (\%)↑ & mTTA (s) & AP (\%) & mTTA (s) & AP (\%) & mTTA (s) \\
\hline
DSA \cite{chan2017anticipating}& 48.1 & 1.34 & 98.7 & 3.08 & 92.3 & 2.95 \\
ACRA \cite{zeng2017agent} & 51.4 & 3.01 & 98.9 & 3.32 & - & - \\
AdaLEA \cite{suzuki2018anticipating}& 52.3 & 3.43 & 99.2 & 3.45 & 92.9 & 3.16 \\
UString \cite{BaoMM2020}& 53.7 & 3.53 & 99.5 & 3.74 & 93.2 & 3.24 \\
DSTA \cite{karim2022dynamic}& 56.1 & 3.66 & 99.3 & 3.87 & 93.5 & 2.87 \\
GSC \cite{wang2023gsc}& 60.4 & 2.55 & 99.4 & 3.68 & 94.9 & 2.62 \\
AccNet \cite{LIAO2024107760}& 60.8 & 3.58 & 99.5 & 3.78 & 95.1 & \underline{3.26} \\
WWW \cite{liao2024when}& 69.2 & \underline{4.26} & \underline{99.7} & 3.93 & \textbf{96.4} & \textbf{3.48} \\
CCAF-Net \cite{liu2025ccaf}& 71.8 & 4.15 & 93.9 & \textbf{4.94} & - & - \\
THAT-NET \cite{liu2023net}& \underline{77.8} & 4.14 & 99.5 & \underline{4.59} & - & - \\
\hline
\textbf{Ours} & \textbf{87.7} & \textbf{4.47} & \textbf{99.7} & 3.70 & \underline{95.4} & 3.18 \\
\hline
\end{tabular}
}

\label{table:balance}
\end{table}

\textbf{Comparison with State-of-the-Art (SOTA).}
Table \ref{table:balance} presents a comparative evaluation of our proposed model against SOTA methods on the DAD, CCD, and A3D datasets. Our model achieves substantial improvements, particularly on the DAD dataset, where it surpasses previous methods by a significant margin, with gains of 12.7\% in AP and 4.9\% in mTTA. Notably, even on the CCD and A3D datasets, where prior methods had approached performance saturation, our model establishes new benchmarks, further solidifying its robustness across diverse accident anticipation tasks. Additionally, as demonstrated in Table \ref{table:bestap}, our model effectively balances  AP and mTTA. It outperforms all baselines even when prioritizing AP alone, achieving an impressive 87.7\% AP, the highest recorded value in this task. Moreover, our model excels in early anticipation, attaining an mTTA of 4.47s, thereby providing an extended response window for accident mitigation. These results emphasize the superior predictive capability and generalization ability of our approach, particularly in real-world driving environments with highly dynamic and uncertain conditions.

\begin{table}[htbp]
\centering
\caption{ {Comparison of models seeking \textbf{best AP} on DAD. \textbf{Bold} and \underline{underlined} values represent the best and second-best performance. Instances where values are not available are marked with a dash (``-'') in each category.}}
 \resizebox{0.8\linewidth}{!}{
\begin{tabular}{cccc}
\hline 
Model & Publication & AP (\%)↑  & mTTA (s) \\
\hline 
ACRA&  ACCV'16 & 51.4 & - \\
DSA&  ACCV'16 & 63.5 & 1.67 \\
UniFormerV2&  ICCV'23 & 65.2 & - \\
VideoSwin&  CVPR'22 & 65.5 & - \\
MVITv2 &  CVPR'21 & 65.5 & - \\
UString&  ACMMM'20 & 68.4 & 1.63 \\
GSC  &  TIV'23 & 68.9 & 1.33 \\
WWW & ACMMM'24 & 69.2 & 4.26 \\
AccNet & AAP'24 & 70.1 & 1.73 \\
CCAF-Net&Neurocomputing'25& 71.8 & 4.15\\
DSTA  & TITS'22 & 72.3 & 1.52 \\
THAT-NET & Inf.Sci'23 & 77.8 & 4.14 \\
\hline
\textbf{Ours} &  - & \textbf{87.7} & \textbf{4.47} \\
\hline
\end{tabular}
}

\label{table:bestap}
\end{table}

\textbf{Comparative Analysis of Model Complexity.}
Table \ref{FLOPs} presents a comparative evaluation of our model's computational efficiency against SOTA methods, highlighting the FLOPs and parameter count. Our model demonstrates remarkable efficiency, utilizing only a fraction of the computational resources required by prior approaches. Specifically, it requires just one-tenth of the FLOPs of the second-most efficient model, DSA, while maintaining a comparable parameter count. This substantial reduction in computational overhead is primarily attributed to our approach, which focuses on scene-level analysis rather than exhaustive object-level relationship modeling.

\begin{table}[htbp]
    \centering
    \caption{ { {Comparisons with SOTA methods in efficiency. \textbf{Bold} and \underline{underlined} values represent the best and second-best performance. Our model has obvious advantages over other advanced models in terms of FLOPs and the number of parameters.
  ``$\downarrow$'' indicates that lower values correspond to improved model performance in each category.}}}
     \resizebox{0.8\linewidth}{!}{
    \begin{tabular}{cccc}
    \hline \specialrule{0em}{1pt}{1pt}
         Method &Backbone& FLOPs (M)↓ & Params (M) \\ \hline
         UniFormerv2 &Transformer& 3600000.00  & 115.00  \\ 
         VideoSwin &Transformer& 282000.00  & 88.10  \\ 
       MVITv2 &Transformer& 206000.00  & 51.00  \\ 
       DSTA &VGG-16& 8868.00  & 4.56  \\ 
       DAA-GNN &VGG-16& 617.22  & \textbf{0.79}  \\ 
       DSA &VGG-16& \underline{421.06}  & 2.10  \\ 
       \hline
       \textbf{Ours}& VGG-16& \textbf{42.90}  & \underline{2.10} \\ \hline
    \end{tabular}
    }

    \label{FLOPs}
\end{table}

\begin{table}[htbp]
\centering
\caption{ {Comparisons with DSTA in inference time (ms). For the ``Simple'' scene, the number of traffic agents is less than 10; which is between 10-15 for the ``Medium'' scene and is greater than 15 for the ``Complex'' scene. \textbf{Bold} values represent the average inference time.}}
 \resizebox{0.7\linewidth}{!}{
\begin{tabular}{ccccc}
\toprule \specialrule{0em}{1pt}{1pt}
              Scene           & \multicolumn{2}{c}{DSTA} & \multicolumn{2}{c}{Ours} \\ \specialrule{0em}{1pt}{1pt} \midrule
              \specialrule{0em}{0.5pt}{0.5pt}
\multirow{3}{*}{Simple}  & 39.37  &                 & 11.32  &                 \\
                         & 39.77  & \textbf{39.83}  & 11.10  & \textbf{11.35}  \\
                         & 40.34  &                 & 11.63  &                 \\ \hline 
\multirow{3}{*}{Medium}  & 43.68  &                 & 11.62  &                 \\ 
                         & 43.54  & \textbf{43.30}   & 11.85  & \textbf{11.65}  \\
                         & 42.69  &                 & 11.49  &               \\   \hline
\multirow{3}{*}{Complex} & 53.37  &                 & 12.03  &                 \\
                         & 47.51  & \textbf{48.48}  & 11.24  & \textbf{11.85}  \\
                         & 44.55  &                 & 12.28  &                 \\ \hline
\multirow{1}{*}{All}&- &\textbf{42.69}  &-  &\textbf{11.58}\\
                       \bottomrule
\end{tabular}
}

\label{inference}
\end{table}

To further assess real-time applicability, we evaluate our model's inference time across traffic scenes of varying complexity. As shown in Table \ref{inference}, our model significantly outperforms DSTA \cite{karim2022dynamic} in terms of inference efficiency. On 1920x1080 resolution videos, our model processes frames in approximately 11.6 ms, which is only 25\% of the DSTA’s inference time. A notable advantage of our approach lies in its resilience to increasing scene complexity. While DSTA exhibits a sharp decline in inference speed as the number of traffic agents increases, our model remains minimally impacted, maintaining near-constant inference times across simple, medium, and complex scenarios. This stability ensures that our model can be deployed in high-density urban environments without significant computational slowdowns.

\begin{figure}[t]
\centering  
\includegraphics[width=0.5\textwidth]{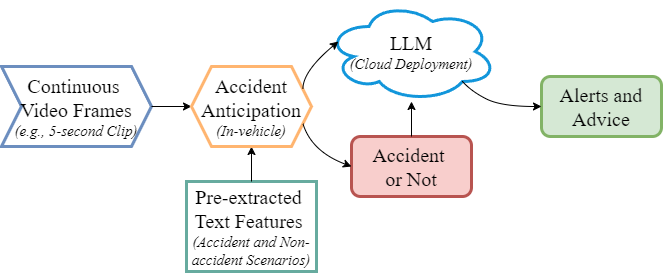}
 
\caption{Deployment framework diagram of the accident anticipation system. The on-board accident anticipation system receives real-time video clips from the dashcam and uses pre-extracted text features to predict accidents. The results are output to the LLM in the cloud, and driving advice is generated through the LLM.}
\label{Fig.7}
\end{figure}

Considering the difference between the actual vehicle computing platform and the current operating environment, the performance of the vehicle platform can be simulated by limiting the computing performance. Taking the Jetson AGX Xavier platform as an example, the GPU frequency is limited to 1.1GHz, the video memory is limited to 3GB, the CPU frequency is limited to 2.2GHz, and the number of cores is limited to 4 cores. At this time, the average inference time of the model is 25.93ms. Due to the challenges of high-performance LLM in-vehicle deployment, in actual deployment scenarios, we recommend the use of an edge-cloud collaborative architecture to place complex LLM processing in the cloud, while the vehicle-mounted end only performs visual feature extraction and real-time accident warning tasks (as Fig. \ref{Fig.7}).

\begin{figure*}[h]
\centering  
\includegraphics[width=1.0\textwidth]{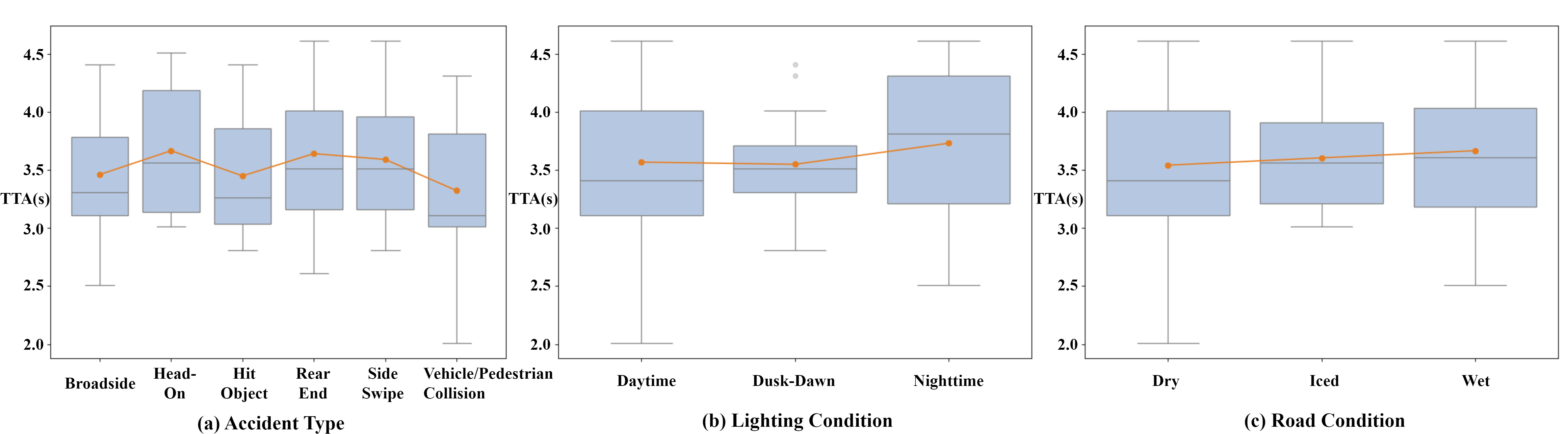}
 
\caption{TTA distribution across accident types, lighting conditions and road conditions. The black box plot shows the distribution and median of TTA, and the orange line shows the mean of TTA.}
\label{Fig.8}
\end{figure*}

\textbf{TTA Pattern Analysis.} To investigate the model's performance across different traffic scenarios, particularly to explore whether there exist consistent patterns between various types of traffic accidents, traffic scenes, and the predicted TTA, we conducted experiments on the CCD dataset. The results are presented in Fig. \ref{Fig.8}.

We first employ Qwen-VL-Max for video understanding, supplemented by manual review to classify each video according to its corresponding traffic scene, lighting condition, and road surface status. Subsequently, we compute the TTA for each traffic scenario and analyze its distribution and mean under various semantic factors related to traffic accidents. Overall, the results reveal noticeable distributional differences and trend patterns in TTA across accident types, lighting conditions, and road surface statuses.

Among different accident types, Rear-End and Head-On collisions exhibit relatively higher median and mean TTA values, indicating that these accidents are generally preceded by more pronounced precursor behaviors, allowing the model to anticipate them earlier. In contrast, Vehicle/Pedestrian Collisions show the lowest TTA values, with distributions heavily concentrated in the low-TTA region, suggesting that such accidents tend to occur more abruptly. The complex and less predictable behaviors of pedestrians make early anticipation significantly more challenging compared to other accident types.

In the experiments analyzing the influence of environmental conditions on TTA, results show that TTA values under Nighttime, Wet, and Iced conditions are slightly higher than those under normal conditions. This observation emphasizes the model's ability to align visual features with pre-loaded textual features under adverse conditions, enabling earlier identification of high-risk factors and extending the lead time for accident warnings.

Overall, these experimental results demonstrate that our model not only achieves higher predictive accuracy compared to previous approaches but also delivers unprecedented inference speed. By integrating domain knowledge through VLMs and adopting a lightweight architecture, we ensure that accident anticipation remains both computationally feasible and scalable. This efficiency makes our model well-suited for deployment in real-world AD systems.

\subsection{Ablation Studies}

\begin{table}[htbp]
\centering
\caption{Ablation study of different modules on the DAD dataset. MA, TF, and LT denote Multi-head Attention, Textual Features, and Learnable Threshold, respectively. ``\( \bullet \)'' represents the reserved modules, while ``\(\times\)'' denotes the corresponding module is removed.}
 \resizebox{0.8\linewidth}{!}{
\begin{tabular}{cccccc}
\hline \specialrule{0em}{1pt}{1pt}
MODEL & MA & TF & LT & AP (\%) & MTTA (s) \\ \hline
A &  \( \times \) & \( \bullet \) & \( \bullet \) & 49.6 & 3.73 \\ 
B & \( \bullet \)&  \( \times \) & \( \bullet \) & 74.3 & 4.46 \\ 
C & \( \bullet \) & \( \bullet \) &  \( \times \) & 61.2 & 4.46 \\
\hline
\textbf{Final} & \( \bullet \)& \( \bullet \)& \( \bullet \) & \textbf{87.7} & \textbf{4.47} \\ \hline
\end{tabular}
}

\label{Ablation}
\end{table}

\begin{table}[htbp]
    \centering
\caption{Ablation study on LLM inputs. ``\( \bullet \)'' represents the LLM receives the corresponding input. In the feedback results, ``\(\checkmark\)'' denotes correct and relevant feedback, while ``\(\times\)'' indicates that the feedback was provided but incorrect.}
    \resizebox{\linewidth}{!}{
    \begin{tabular}{cccccccc}
    \hline
    \specialrule{0em}{2pt}{2pt}
        \multirow{2}{*}{} & \multirow{2}{*}{\makecell[c]{Video \\ Frame}} & \multirow{2}{*}{\makecell[c]{Accident\\ Anticipation}}& \multirow{2}{*}{\makecell[c]{Domain \\Knowledge}} & \multicolumn{4}{c}{\textbf{LLM Feedback}} \\ 
        \cmidrule(lr){5-8}
        &           ~            &        ~     &       ~     & Alert & Scene & Type &  Advice \\ \hline
        A & \( \bullet \) & ~ & ~ &  \( \times \) &  \( \times \) &  \( \times \) &  \( \times \)\\ 
        B & \( \bullet \) & \( \bullet \) & ~ & \checkmark &  \( \times \) &  \( \times \) &  \( \times \) \\
   
        \textbf{Ours} & \( \bullet \) & \( \bullet \) & \( \bullet \) & \checkmark & \checkmark & \checkmark & \checkmark \\ \hline
    \end{tabular}
    }
   
    \label{feedback}
\end{table}

\begin{figure*}[t]
\centering  
\includegraphics[width=\textwidth]{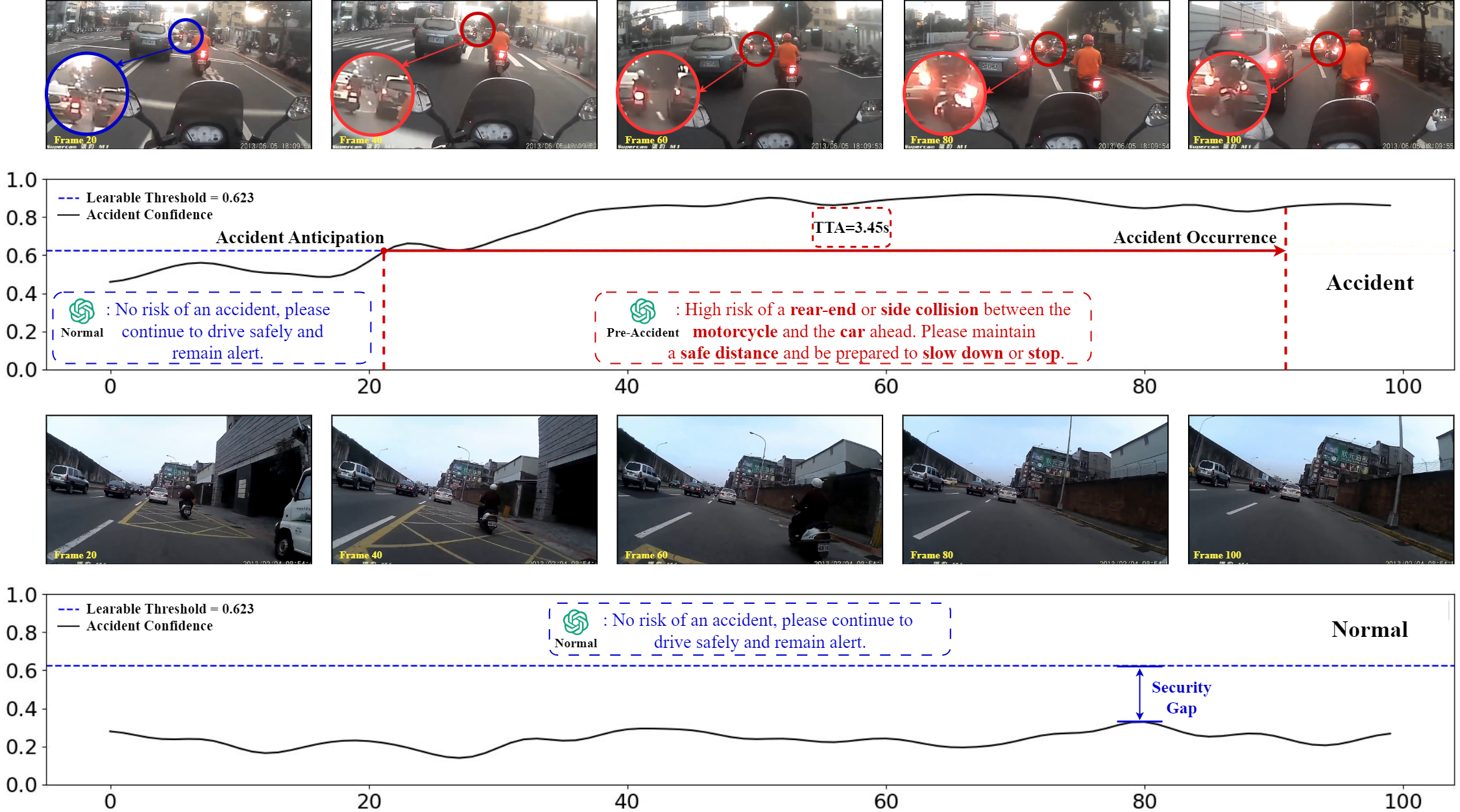}

\caption{{Visualization of our model's performance on the DAD dataset. A representative positive and negative example are selected to illustrate the model's predictive capabilities. It accurately anticipates the occurrence of an accident, analyzes potential contributing factors, and provides context-aware driving recommendations, demonstrating its effectiveness in real-world scenes.}}
\label{Fig.9}
\end{figure*}

Table \ref{Ablation} presents the ablation results on the DAD dataset, evaluating the contributions of the textual branch, temporal attention module, and learnable threshold to overall model performance. 
Replacing the temporal attention module with a fully connected (FC) layer (Model A) leads to a significant drop in both AP and mTTA, highlighting the critical role of temporal learning in capturing dynamic accident-prone scenarios. Removing the textual branch (Model B) also results in a noticeable decline in AP, demonstrating that incorporating domain knowledge from structured accident reports improves accident anticipation accuracy. Additionally, using a fixed threshold of 0.5 (Model C) increases false positive rates, ultimately reducing the model's reliability in distinguishing high-risk situations from non-accidents.

Correspondingly, Table \ref{feedback} presents the impact of accident anticipation results and domain knowledge on LLM-generated feedback. The results emphasize that domain knowledge enhances the LLM's ability to focus on accident-related factors, providing more precise and contextually relevant analysis. To further investigate our findings, we conduct a set of controlled experiments, which can be summarized as follows:

 \textbf{Experiment A:} The LLM receives only the current video frame as input and must determine whether to issue an accident warning and provide driving recommendations based solely on visual content. Under this condition, the LLM struggles to assess accident probability accurately, fails to identify key accident-related factors in the scene, and is unable to generate meaningful recommendations.
    
 \textbf{Experiment B:} Both the current video frame and the model's accident anticipation results are provided as input to the LLM. While the LLM can now correctly decide whether to issue a warning based on the model's predictions, it lacks a structured understanding of the contributing factors. As a result, it generates only generic driving advice, without offering specific, scenario-adaptive recommendations.
    
 \textbf{Ours:} We input the video frame, accident anticipation results, and domain knowledge into the LLM. This structured approach directs the LLM's attention toward causal accident factors, enabling it to reason more effectively about the scenario. Consequently, the LLM can generate targeted and actionable outputs, including accurate accident warnings, descriptions of critical factors in the scene, predictions of accident types, and corresponding driving recommendations tailored to the situation.

To sum up, these findings demonstrate the effectiveness of integrating domain knowledge into accident anticipation, allowing the model to bridge the gap between statistical learning and real-world reasoning. By incorporating structured accident reports and leveraging multimodal learning, our approach ensures both high prediction accuracy and interpretable decision-making.

\begin{figure*}[t]
\centering  
\includegraphics[width=0.85\textwidth]{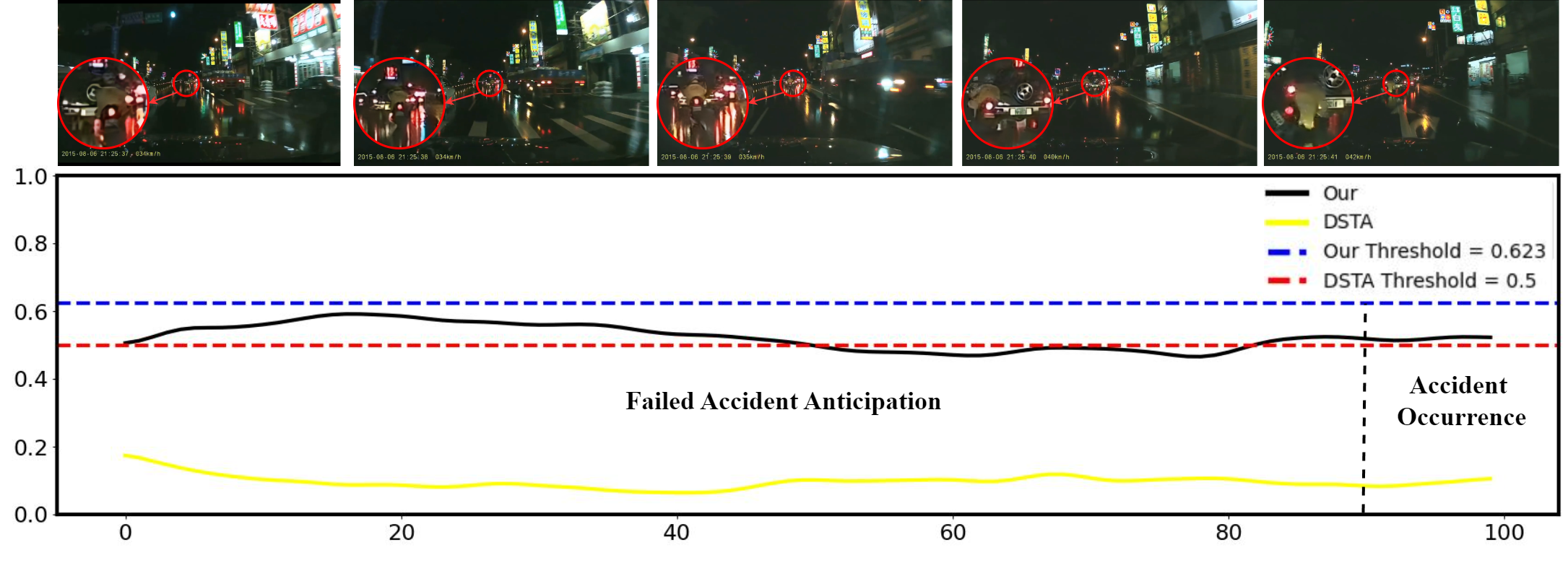}

\caption{{Visualization of a failed accident anticipation case and comparison with the DSTA model output. }In this scenario, the accident anticipation model fails to predict the accident accurately due to poor lighting conditions and the significant distance between the accident participants and the ego vehicle, leading to minimal valid visual information for analysis.}
\label{Fig.10}
\end{figure*}

\subsection{Qualitative Analysis} 
Fig. \ref{Fig.9} presents qualitative visualizations of accident anticipation results using samples from the DAD testing dataset. For the first traffic accident scenario, during the early phase of the video, the accident confidence score predicted by our model fluctuates slightly below the threshold but remains at a relatively high level due to the complexity of the traffic environment. As the motorcycle in front attempts to overtake and the silver car initiates a lane change (highlighted in the figure), the system detects these high-risk interactions, leading to a rapid increase in accident confidence. Once the confidence surpasses the threshold, the model issues an accident warning, analyzes the current scene, and leverages an LLM to provide actionable driving recommendations. This demonstrates the model's ability to recognize critical interactions between traffic participants and dynamically adjust risk assessments. For the second non-accident scenario, our model consistently maintains a low accident confidence score with normal fluctuations. Unlike traditional methods that set a fixed accident confidence threshold at 0.5, our approach employs a learnable threshold, which provides a more stringent accident classification criterion. This adaptive thresholding enhances the safety margin in non-accident cases, significantly reducing false alarms while maintaining high sensitivity to genuine risks.

Besides, we analyze a failure case to examine the limitations of our model. As shown in Fig. \ref{Fig.10}, under extremely poor visual conditions, the system assigns a high accident confidence score based on adverse environmental factors but fails to detect key vehicle behaviors, preventing it from crossing the threshold to issue a warning. This highlights a fundamental limitation—reliance on visual input quality. When image clarity is compromised due to low lighting or occlusions, the model loses essential visual cues, leading to prediction failures. In addition, we further compare this failure case with DSTA, the results reveal that DSTA, which relies solely on visual input, also fails to anticipate the accident in this scenario. Notably, because DSTA does not incorporate environmental factors or domain knowledge, its confidence scores are significantly lower than those of our model. This comparison underscores the advantage of integrating environmental context and multimodal reasoning in anticipation, demonstrating the robustness of our approach while also highlighting areas for future improvement.

\begin{figure*}[h]
\centering  
\includegraphics[width=0.85\textwidth]{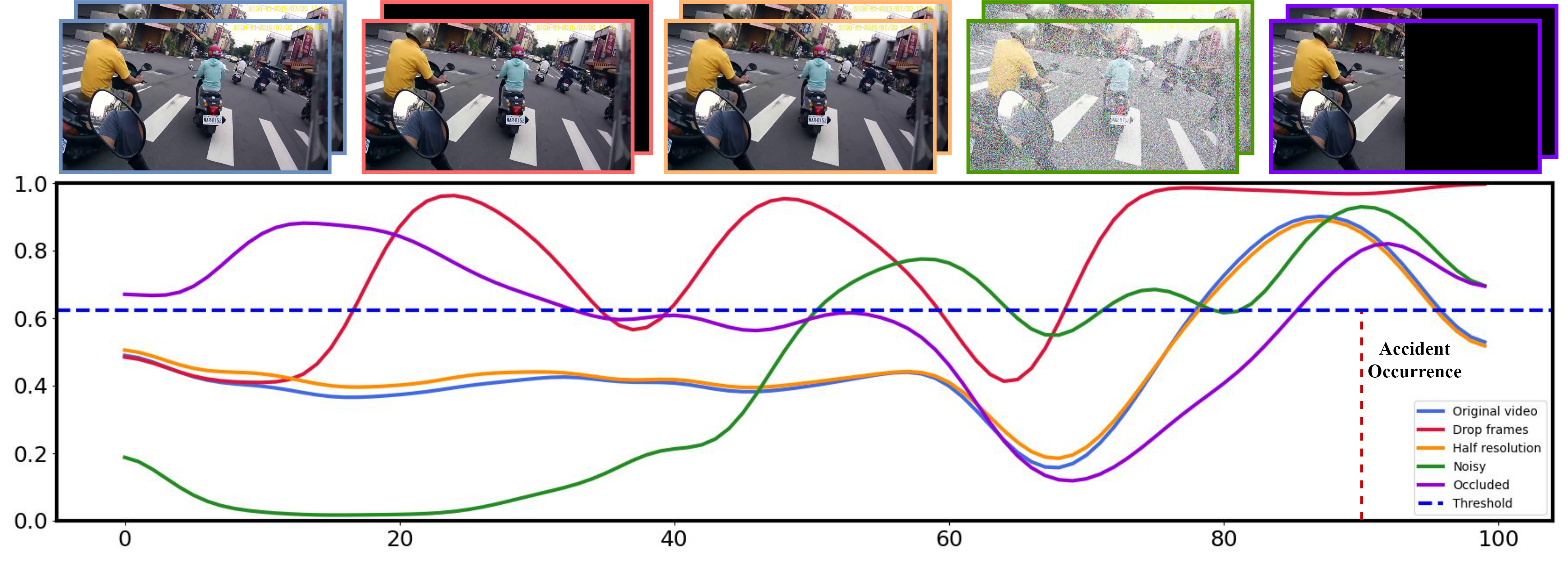}
 
\caption{Robustness analysis considering real-world challenges. The top image shows the visual characteristics of the input video after processing in different experiments, and the bottom image shows the accident confidence output by these experiments.}
\label{Fig.11}
\end{figure*}

To address common real-world challenges, we take a positive case as an example to explore the robustness of our model and the possible impact of common sensor failures. Fig. \ref{Fig.11} shows the accident confidence output of our model under different input constraints. In this experiment, we simulate the following common sensor failures by processing the original video.

\begin{itemize}

\item \textbf{Drop frames.} Short-term sensor failure or errors in the data transmission process will cause the input video fragment to be lost. We periodically delete 10 frames (0.5 seconds) from the original video to simulate this situation.

\item \textbf{Half resolution.} We reduce the resolution of the original video by 50\% to analyze the performance of the model under low-resolution input.

\item \textbf{Noisy.} In order to simulate the situation where the sensor is affected and the input video is blurred, we add Gaussian noise with a mean of 0 and a standard deviation of 25 to the original video.

\item \textbf{Occluded.} In the real world, there are often video occlusion problems, such as the dash cam being partially blocked by debris. Therefore, we occluded the right half of the original video to observe the performance of the model for video occlusion.
\end{itemize}

Experimental results show that our model can cope with the challenges brought by these real-world sensor failures and output correct accident anticipation results. However, sensor problems also bring some challenges to the accident anticipation system.

First, in low-resolution mode, the output of the model is similar to that in normal situations, proving that changes in video resolution have little impact on accident anticipation. In addition, when noise is added, the model's prediction results are relatively stable. However, in the case of video frame loss and large-area occlusion, since the performance of image loss is similar to the loss of dashcam images due to vehicle accidents in some scenarios, the model will more easily predict the current scene as a possible accident. This conclusion can be confirmed by the performance of the "Drop frames" experiment. When the video frame is periodically obscured, the accident confidence output by the model also fluctuates periodically.

\section{Conclusion}
\label{Conclusion}

In this study, we present an efficient dual-branch and multimodal learning traffic accident anticipation model that integrates domain knowledge with Vision-Language Models. Additionally, we introduce Large Language Models to generate verbal accident alerts, enhancing human-machine interaction and interpretability. Notably, we are the first to incorporate structured traffic accident reports into accident anticipation, leveraging this underutilized yet valuable data source to improve both efficiency and accuracy. By systematically integrating domain knowledge, our model simplifies the accident anticipation process while achieving state-of-the-art performance across key metrics on real-world datasets, including DAD, CCD, and A3D. Our research introduces accident reports, a huge but underutilized data, into accident anticipation, which has far-reaching significance for future autonomous driving technology. This study demonstrates the correlation between text and video data and opens up a new avenue for the study of accident anticipation in autonomous driving. Future research can further explore the potential link between accident reports and accident videos, transforming the huge amount of textual data into the much needed visual data with rich annotations, and then support the training of autonomous driving systems. In terms of accident anticipation, in addition to whether an accident will occur, future research can further refine the scene and provide rich driving suggestions based on different scenes.

\section*{CODE availability}

The proposed code of this study is available on ETS data platform (ID: ETS-Data-25-00023).


%

\section*{Acknowledgment}
This work was supported by the Science and Technology Development Fund of Macau [0122/2024/RIB2, 0215/2024/AGJ, 001/2024/SKL], the Research Services and Knowledge Transfer Office, University of Macau [SRG2023-00037-IOTSC, MYRG-GRG2024-00284-IOTSC], the Shenzhen-Hong Kong-Macau Science and Technology Program Category C [SGDX20230821095159012], the State Key Lab of Intelligent Transportation System [2024-B001], and the Jiangsu Provincial Science and Technology Program [BZ2024055].



\printcredits


\bibliography{cas-refs}

\begin{thebibliography}{}

\bibitem[Abdel-Aty and Ding, 2024]{abdel2024matched}
Abdel-Aty, M. and Ding, S. (2024).
\newblock A matched case-control analysis of autonomous vs human-driven vehicle accidents.
\newblock {\em Nature Communications}, 15(1):4931.

\bibitem[Achiam et~al., 2023]{achiam2023gpt}
Achiam, J., Adler, S., Agarwal, S., Ahmad, L., Akkaya, I., Aleman, F.~L., Almeida, D., Altenschmidt, J., Altman, S., Anadkat, S., et~al. (2023).
\newblock Gpt-4 technical report.
\newblock {\em arXiv preprint arXiv:2303.08774}.

\bibitem[Ahmed et~al., 2023]{ahmed2023road}
Ahmed, S.~K., Mohammed, M.~G., Abdulqadir, S.~O., El-Kader, R. G.~A., El-Shall, N.~A., Chandran, D., Rehman, M. E.~U., and Dhama, K. (2023).
\newblock Road traffic accidental injuries and deaths: A neglected global health issue.
\newblock {\em Health science reports}, 6(5):e1240.

\bibitem[Bao et~al., 2020]{BaoMM2020}
Bao, W., Yu, Q., and Kong, Y. (2020).
\newblock Uncertainty-based traffic accident anticipation with spatio-temporal relational learning.
\newblock In {\em ACM Multimedia Conference}.

\bibitem[Bao et~al., 2021]{bao2021drive}
Bao, W., Yu, Q., and Kong, Y. (2021).
\newblock Drive: Deep reinforced accident anticipation with visual explanation.
\newblock In {\em Proceedings of the IEEE/CVF International Conference on Computer Vision}, pages 7619--7628.

\bibitem[Chan et~al., 2017]{chan2017anticipating}
Chan, F.-H., Chen, Y.-T., Xiang, Y., and Sun, M. (2017).
\newblock Anticipating accidents in dashcam videos.
\newblock In {\em Computer Vision--ACCV 2016: 13th Asian Conference on Computer Vision, Taipei, Taiwan, November 20-24, 2016, Revised Selected Papers, Part IV 13}, pages 136--153. Springer.

\bibitem[Chen et~al., 2022]{chen2022review}
Chen, J., Wang, Q., Cheng, H.~H., Peng, W., and Xu, W. (2022).
\newblock A review of vision-based traffic semantic understanding in itss.
\newblock {\em IEEE Transactions on Intelligent Transportation Systems}.

\bibitem[Collingwood, 2017]{collingwood2017privacy}
Collingwood, L. (2017).
\newblock Privacy implications and liability issues of autonomous vehicles.
\newblock {\em Information \& Communications Technology Law}, 26(1):32--45.

\bibitem[Da et~al., 2024]{da2024open}
Da, L., Liou, K., Chen, T., Zhou, X., Luo, X., Yang, Y., and Wei, H. (2024).
\newblock Open-ti: Open traffic intelligence with augmented language model.
\newblock {\em International Journal of Machine Learning and Cybernetics}, pages 1--26.

\bibitem[Dan et~al., 2012]{dan2012multi}
Dan, C., Ueli, M., Jonathan, M., and J{\"u}rgen, S.-h. (2012).
\newblock Multi-column deep neural network for traffic sign classification.
\newblock {\em Neural networks}, 32(1):333--338.

\bibitem[Dash et~al., 2022]{dash2022review}
Dash, T., Chitlangia, S., Ahuja, A., and Srinivasan, A. (2022).
\newblock A review of some techniques for inclusion of domain-knowledge into deep neural networks.
\newblock {\em Scientific Reports}, 12(1):1040.

\bibitem[Devlin et~al., 2019]{devlin2019bert}
Devlin, J., Chang, M.-W., Lee, K., and Toutanova, K. (2019).
\newblock Bert: Pre-training of deep bidirectional transformers for language understanding.
\newblock In {\em Proceedings of the 2019 conference of the North American chapter of the association for computational linguistics: human language technologies, volume 1 (long and short papers)}, pages 4171--4186.

\bibitem[Fang et~al., 2022]{fang2022traffic}
Fang, J., Qiao, J., Bai, J., Yu, H., and Xue, J. (2022).
\newblock Traffic accident detection via self-supervised consistency learning in driving scenarios.
\newblock {\em IEEE Transactions on Intelligent Transportation Systems}, 23(7):9601--9614.

\bibitem[Fang et~al., 2023]{fang2023vision}
Fang, J., Qiao, J., Xue, J., and Li, Z. (2023).
\newblock Vision-based traffic accident detection and anticipation: A survey.
\newblock {\em IEEE Transactions on Circuits and Systems for Video Technology}.

\bibitem[Fang et~al., 2019]{fang2019dada}
Fang, J., Yan, D., Qiao, J., Xue, J., Wang, H., and Li, S. (2019).
\newblock Dada-2000: Can driving accident be predicted by driver attentionƒ analyzed by a benchmark.
\newblock In {\em 2019 IEEE Intelligent Transportation Systems Conference (ITSC)}, pages 4303--4309. IEEE.

\bibitem[Fang et~al., 2021]{fang2021dada}
Fang, J., Yan, D., Qiao, J., Xue, J., and Yu, H. (2021).
\newblock Dada: Driver attention prediction in driving accident scenarios.
\newblock {\em IEEE transactions on intelligent transportation systems}, 23(6):4959--4971.

\bibitem[Fatima et~al., 2021]{fatima2021global}
Fatima, M., Khan, M. U.~K., and Kyung, C.-M. (2021).
\newblock Global feature aggregation for accident anticipation.
\newblock In {\em 2020 25th International Conference on Pattern Recognition (ICPR)}, pages 2809--2816. IEEE.

\bibitem[GDPR, 2018]{gdpr2018general}
GDPR, E. (2018).
\newblock General data protection regulation (gdpr).

\bibitem[Karim et~al., 2022]{karim2022dynamic}
Karim, M.~M., Li, Y., Qin, R., and Yin, Z. (2022).
\newblock A dynamic spatial-temporal attention network for early anticipation of traffic accidents.
\newblock {\em IEEE Transactions on Intelligent Transportation Systems}, 23(7):9590--9600.

\bibitem[Kumar et~al., 2020]{kumar2020iot}
Kumar, N., Acharya, D., and Lohani, D. (2020).
\newblock An iot-based vehicle accident detection and classification system using sensor fusion.
\newblock {\em IEEE Internet of Things Journal}, 8(2):869--880.

\bibitem[Le et~al., 2020]{le2020attention}
Le, T.-N., Ono, S., Sugimoto, A., and Kawasaki, H. (2020).
\newblock Attention r-cnn for accident detection.
\newblock In {\em 2020 IEEE intelligent vehicles symposium (IV)}, pages 313--320. IEEE.

\bibitem[Li et~al., 2022]{li2022network}
Li, J., Xie, N., Zhang, K., Guo, F., Hu, S., and Chen, X.~M. (2022).
\newblock Network-scale traffic prediction via knowledge transfer and regional mfd analysis.
\newblock {\em Transportation research part C: emerging technologies}, 141:103719.

\bibitem[Li et~al., 2024]{li2024cognitive}
Li, L.-L., Fang, J., and Xue, J. (2024).
\newblock Cognitive traffic accident anticipation.
\newblock {\em IEEE Intelligent Transportation Systems Magazine}.

\bibitem[Liao et~al., 2024a]{liao2024real}
Liao, H., Li, Y., Li, Z., Bian, Z., Lee, J., Cui, Z., Zhang, G., and Xu, C. (2024a).
\newblock Real-time accident anticipation for autonomous driving through monocular depth-enhanced 3d modeling.
\newblock {\em Accident Analysis \& Prevention}, 207:107760.

\bibitem[Liao et~al., 2024b]{LIAO2024107760}
Liao, H., Li, Y., Li, Z., Bian, Z., Lee, J., Cui, Z., Zhang, G., and Xu, C. (2024b).
\newblock Real-time accident anticipation for autonomous driving through monocular depth-enhanced 3d modeling.
\newblock {\em Accident Analysis \& Prevention}, 207:107760.

\bibitem[Liao et~al., 2024c]{liao2024when}
Liao, H., Li, Y., Li, Z., Wang, C., Guan, Y., Tam, K., Tian, C., Li, L., and zhong Xu, C. (2024c).
\newblock When, where, and what? a benchmark for accident anticipation and localization with large language models.
\newblock In {\em ACM Multimedia 2024}.

\bibitem[Liao et~al., 2024d]{liao2024gpt}
Liao, H., Shen, H., Li, Z., Wang, C., Li, G., Bie, Y., and Xu, C. (2024d).
\newblock Gpt-4 enhanced multimodal grounding for autonomous driving: Leveraging cross-modal attention with large language models.
\newblock {\em Communications in Transportation Research}, 4:100116.

\bibitem[Liu et~al., 2025]{liu2025ccaf}
Liu, W., Li, Y., Zhang, T., Gao, Y., Wei, L., and Chen, J. (2025).
\newblock Ccaf-net: Cascade complementarity-aware fusion network for traffic accident prediction in dashcam videos.
\newblock {\em Neurocomputing}, 624:129285.

\bibitem[Liu et~al., 2023]{liu2023net}
Liu, W., Zhang, T., Lu, Y., Chen, J., and Wei, L. (2023).
\newblock That-net: Two-layer hidden state aggregation based two-stream network for traffic accident prediction.
\newblock {\em Information Sciences}, 634:744--760.

\bibitem[Mishra et~al., 2023]{mishra2023sensing}
Mishra, S., Rajendran, P.~K., Vecchietti, L.~F., and Har, D. (2023).
\newblock Sensing accident-prone features in urban scenes for proactive driving and accident prevention.
\newblock {\em IEEE Transactions on Intelligent Transportation Systems}, 24(9):9401--9414.

\bibitem[Nie et~al., 2025]{nie2025data}
Nie, J., Jiang, J., Li, Y., Wang, H., Ercisli, S., and Lv, L. (2025).
\newblock Data and domain knowledge dual-driven artificial intelligence: Survey, applications, and challenges.
\newblock {\em Expert Systems}, 42(1):e13425.

\bibitem[Parsa et~al., 2019]{parsa2019real}
Parsa, A.~B., Taghipour, H., Derrible, S., and Mohammadian, A.~K. (2019).
\newblock Real-time accident detection: Coping with imbalanced data.
\newblock {\em Accident Analysis \& Prevention}, 129:202--210.

\bibitem[Radford et~al., 2021]{radford2021learning}
Radford, A., Kim, J.~W., Hallacy, C., Ramesh, A., Goh, G., Agarwal, S., Sastry, G., Askell, A., Mishkin, P., Clark, J., et~al. (2021).
\newblock Learning transferable visual models from natural language supervision.
\newblock In {\em International conference on machine learning}, pages 8748--8763. PMLR.

\bibitem[Radford et~al., 2018]{radford2018improving}
Radford, A., Narasimhan, K., Salimans, T., Sutskever, I., et~al. (2018).
\newblock Improving language understanding by generative pre-training.

\bibitem[Simonyan and Zisserman, 2014]{simonyan2014very}
Simonyan, K. and Zisserman, A. (2014).
\newblock Very deep convolutional networks for large-scale image recognition.
\newblock {\em arXiv preprint arXiv:1409.1556}.

\bibitem[Simonyan and Zisserman, 2015]{Simonyan15}
Simonyan, K. and Zisserman, A. (2015).
\newblock Very deep convolutional networks for large-scale image recognition.
\newblock In {\em International Conference on Learning Representations}.

\bibitem[Song et~al., 2024]{song2024dynamic}
Song, W., Li, S., Chang, T., Xie, K., Hao, A., and Qin, H. (2024).
\newblock Dynamic attention augmented graph network for video accident anticipation.
\newblock {\em Pattern Recognition}, 147:110071.

\bibitem[Suzuki et~al., 2018]{suzuki2018anticipating}
Suzuki, T., Kataoka, H., Aoki, Y., and Satoh, Y. (2018).
\newblock Anticipating traffic accidents with adaptive loss and large-scale incident db.
\newblock In {\em Proceedings of the IEEE conference on computer vision and pattern recognition}, pages 3521--3529.

\bibitem[Thakur et~al., 2024]{thakur2024graph}
Thakur, N., Gouripeddi, P., and Li, B. (2024).
\newblock Graph (graph): A nested graph-based framework for early accident anticipation.
\newblock In {\em Proceedings of the IEEE/CVF Winter Conference on Applications of Computer Vision}, pages 7533--7541.

\bibitem[Vaswani et~al., 2017]{vaswani2017attention}
Vaswani, A., Shazeer, N., Parmar, N., Uszkoreit, J., Jones, L., Gomez, A.~N., Kaiser, {\L}., and Polosukhin, I. (2017).
\newblock Attention is all you need.
\newblock {\em Advances in neural information processing systems}, 30.

\bibitem[Wang et~al., 2024]{wang2024accidentgpt}
Wang, L., Ren, Y., Jiang, H., Cai, P., Fu, D., Wang, T., Cui, Z., Yu, H., Wang, X., Zhou, H., et~al. (2024).
\newblock Accidentgpt: A v2x environmental perception multi-modal large model for accident analysis and prevention.
\newblock In {\em 2024 IEEE Intelligent Vehicles Symposium (IV)}, pages 472--477. IEEE.

\bibitem[Wang et~al., 2023]{wang2023gsc}
Wang, T., Chen, K., Chen, G., Li, B., Li, Z., Liu, Z., and Jiang, C. (2023).
\newblock Gsc: A graph and spatio-temporal continuity based framework for accident anticipation.
\newblock {\em IEEE Transactions on Intelligent Vehicles}, 9(1):2249--2261.

\bibitem[Wu et~al., 2024]{wu2024vadclip}
Wu, P., Zhou, X., Pang, G., Zhou, L., Yan, Q., Wang, P., and Zhang, Y. (2024).
\newblock Vadclip: Adapting vision-language models for weakly supervised video anomaly detection.
\newblock In {\em Proceedings of the AAAI Conference on Artificial Intelligence}, volume~38, pages 6074--6082.

\bibitem[Xu et~al., 2022]{xu2022trajectory}
Xu, X., Liu, W., and Yu, L. (2022).
\newblock Trajectory prediction for heterogeneous traffic-agents using knowledge correction data-driven model.
\newblock {\em Information Sciences}, 608:375--391.

\bibitem[Yao et~al., 2019]{yao2019unsupervised}
Yao, Y., Xu, M., Wang, Y., Crandall, D.~J., and Atkins, E.~M. (2019).
\newblock Unsupervised traffic accident detection in first-person videos.
\newblock In {\em 2019 IEEE/RSJ International Conference on Intelligent Robots and Systems (IROS)}, pages 273--280. IEEE.

\bibitem[Yu et~al., 2020]{yu2020bdd100k}
Yu, F., Chen, H., Wang, X., Xian, W., Chen, Y., Liu, F., Madhavan, V., and Darrell, T. (2020).
\newblock Bdd100k: A diverse driving dataset for heterogeneous multitask learning.
\newblock In {\em Proceedings of the IEEE/CVF conference on computer vision and pattern recognition}, pages 2636--2645.

\bibitem[Yuan et~al., 2024]{yuan2024unist}
Yuan, Y., Ding, J., Feng, J., Jin, D., and Li, Y. (2024).
\newblock Unist: A prompt-empowered universal model for urban spatio-temporal prediction.
\newblock In {\em Proceedings of the 30th ACM SIGKDD Conference on Knowledge Discovery and Data Mining}, pages 4095--4106.

\bibitem[Zeng et~al., 2017]{zeng2017agent}
Zeng, K.-H., Chou, S.-H., Chan, F.-H., Carlos~Niebles, J., and Sun, M. (2017).
\newblock Agent-centric risk assessment: Accident anticipation and risky region localization.
\newblock In {\em Proceedings of the IEEE Conference on Computer Vision and Pattern Recognition}, pages 2222--2230.

\bibitem[Zhang et~al., 2024]{zhang2024long}
Zhang, B., Zhang, P., Dong, X., Zang, Y., and Wang, J. (2024).
\newblock Long-clip: Unlocking the long-text capability of clip.
\newblock {\em arXiv preprint arXiv:2403.15378}.

\bibitem[Zhang et~al., 2019]{zhang2019graph}
Zhang, S., Tong, H., Xu, J., and Maciejewski, R. (2019).
\newblock Graph convolutional networks: a comprehensive review.
\newblock {\em Computational Social Networks}, 6(1):1--23.

\bibitem[Zhao et~al., 2019]{zhao2019t}
Zhao, L., Song, Y., Zhang, C., Liu, Y., Wang, P., Lin, T., Deng, M., and Li, H. (2019).
\newblock T-gcn: A temporal graph convolutional network for traffic prediction.
\newblock {\em IEEE transactions on intelligent transportation systems}, 21(9):3848--3858.

\bibitem[Zhou and Knoll, 2024]{zhou2024gpt}
Zhou, X. and Knoll, A.~C. (2024).
\newblock Gpt-4v as traffic assistant: An in-depth look at vision language model on complex traffic events.
\newblock {\em arXiv preprint arXiv:2402.02205}.

\end{thebibliography}

\newpage
\bio{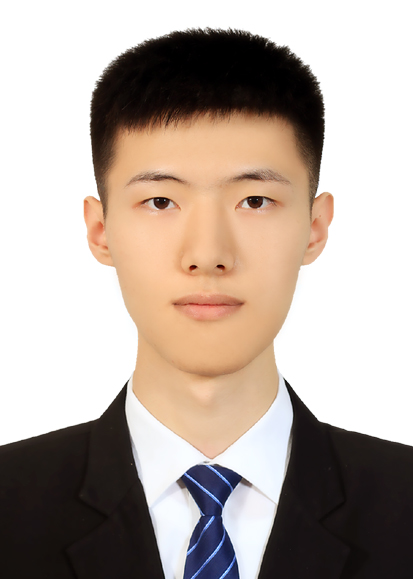} Yanchen Guan is currently a PhD student at the State Key Laboratory of Internet of Things for Smart City and the Department of Civil Engineering at the University of Macau. He holds a master’s degree in Mobility Engineering from Politecnico di Milano (2023) and a bachelor’s degree in Mechatronics Engineering from Harbin Institute of Technology (2019). His research primarily focuses on autonomous driving, intelligent transportation systems, mechanical structures, and data analysis.
\endbio

\bio{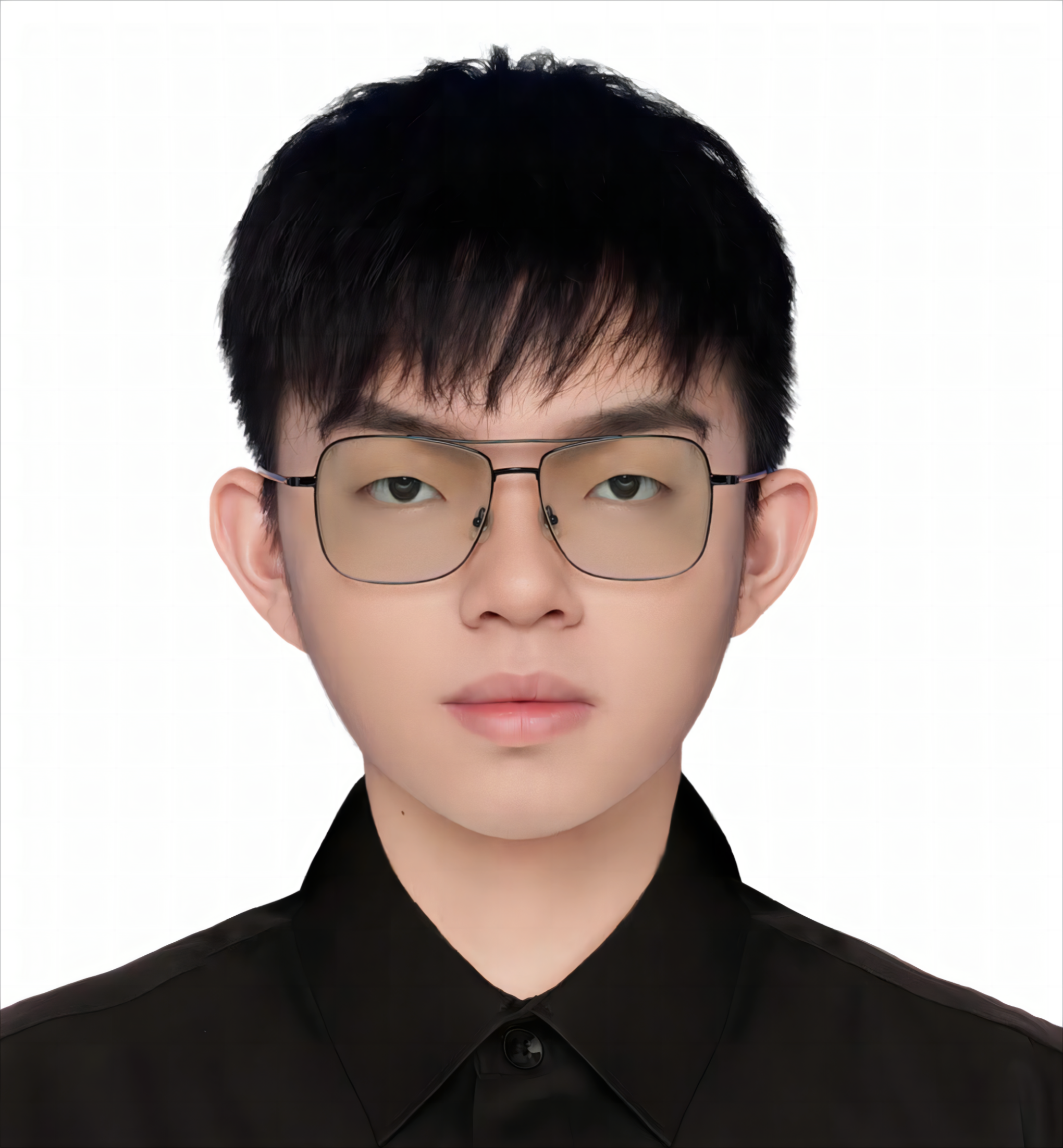} Haicheng Liao received the B.S. degree in software engineering from the University of Electronic Science and Technology of China (UESTC) in 2022. He is currently pursuing the Ph.D. degree at the State Key Laboratory of Internet of Things for Smart City and the Department of Computer and Information Science, University of Macau. His research interests include connected autonomous vehicles and the application of deep reinforcement learning to autonomous driving.
\endbio

\bio{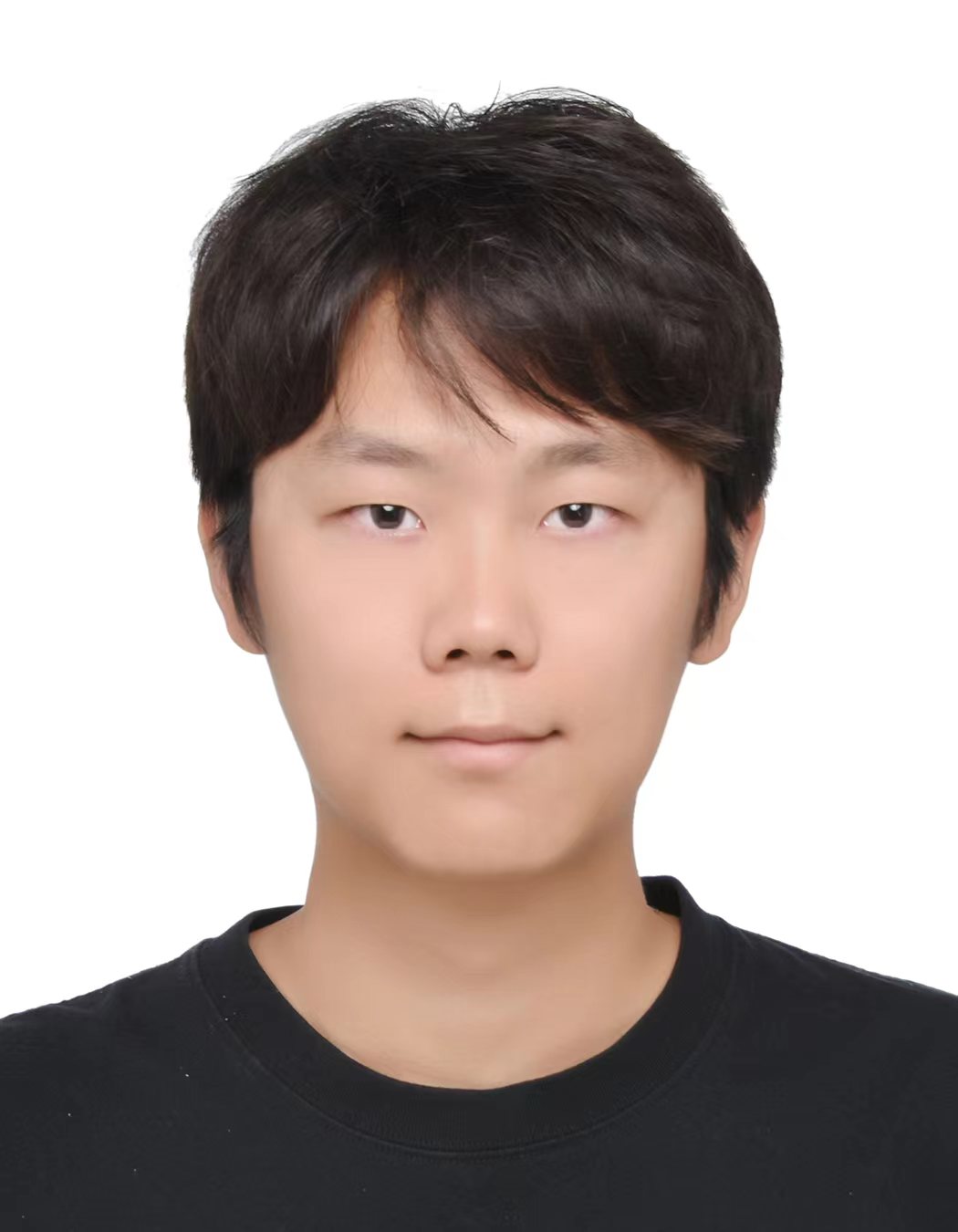} 
Chengyue Wang is currently pursuing a Ph.D. degree at the State Key Laboratory of Internet of Things for Smart City and the Department of Civil Engineering, University of Macau. He received his M.S. degree in civil engineering from the University of Illinois Urbana-Champaign (UIUC) in 2022. He received his B.E. degree in transportation engineering from Chang'an University in 2021. His research interests include connected autonomous vehicles and the application of deep reinforcement learning to autonomous driving.
\endbio

\bio{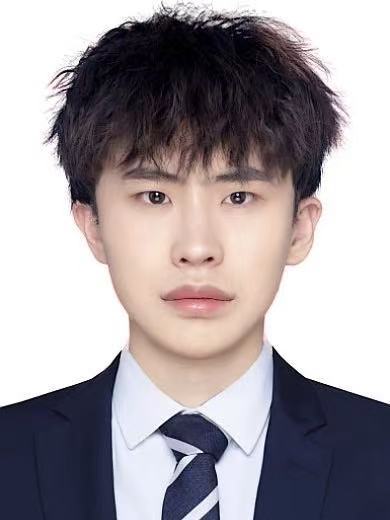} Bonan Wang is currently pursuing an M.S. degree at the State Key Laboratory of Internet of Things for Smart City and the Department of Computer and Information Science, University of Macau. He received a B.S. degree in Data Science and Big Data Technology from Shaanxi University of Science \& Technology in 2023. His research interests include trajectory prediction for autonomous driving.
\endbio


\bio{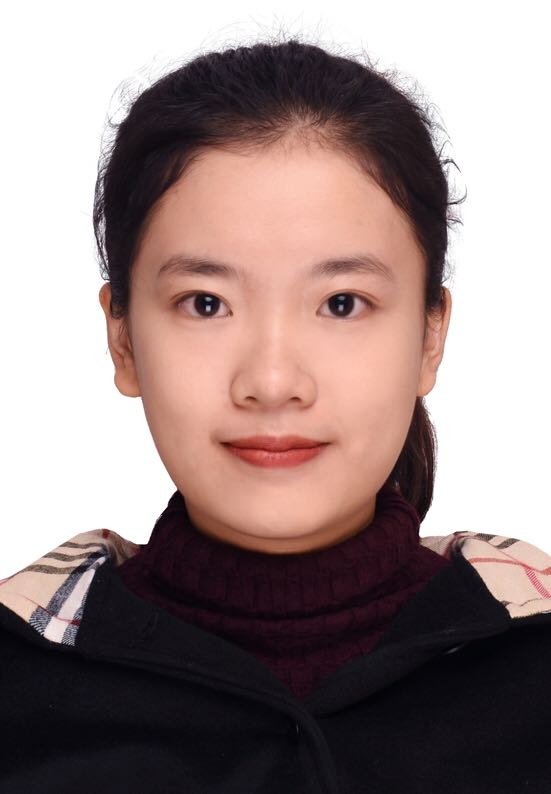}
Jiaxun Zhang is currently a pursuing Ph.D. degree
at the State Key Laboratory of Internet of Things
for Smart City and the Department of Civil Engineering at the University of Macau. She received an
MS degree in Integrated Sustainable Design from
National University of Singapore (NUS) in 2023 and
a B.S. degree in Traffic engineering from the South
China University of Technology (SCUT) in 2022.
Her research interests include the application of deep
reinforcement learning to autonomous driving and
intelligent transportation systems.
\endbio
\newpage

\bio{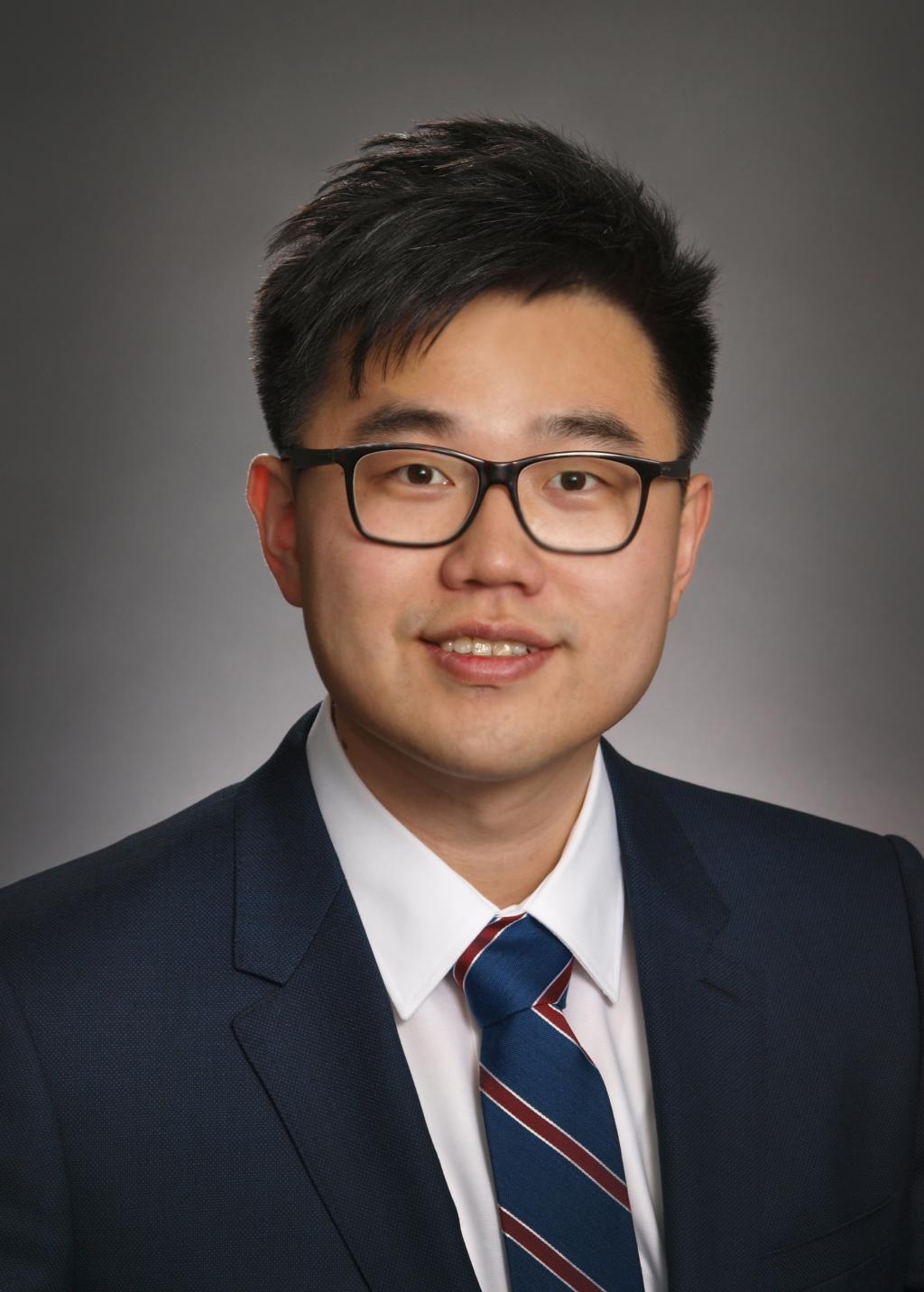}
Jia Hu works as a ZhongTe Distinguished Chair in Cooperative Automation in the College of Transportation Engineering at Tongji University. Before joining Tongji, he was a research associate at the Federal Highway Administration, USA (FHWA). He is an Associate Editor of the IEEE Transaction on Intelligent Transportation Systems, IEEE Transaction on Intelligent Vehicle, American Society of Civil Engineers Journal of Transportation Engineering, IEEE Open Journal in Intelligent Transportation Systems, an assistant editor of the Journal of Intelligent Transportation Systems, an advisory editorial board member for the Transportation Research Part C, an associate editor for IEEE Intelligent Vehicles Symposium since 2018, and an associate editor for IEEE Intelligent Transportation Systems Conference since 2019. Furthermore, he is a member of the Board of Governor of IEEE ITS Society, TRB (a division of the National Academies) Vehicle Highway Automation Committee, Freeway Operation Committee and Simulation subcommittee of Traffic Signal Systems Committee, and a member of CAV Impact Committee and Artificial Intelligence Committee of ASCE Transportation and Development Institute.
\endbio

\bio{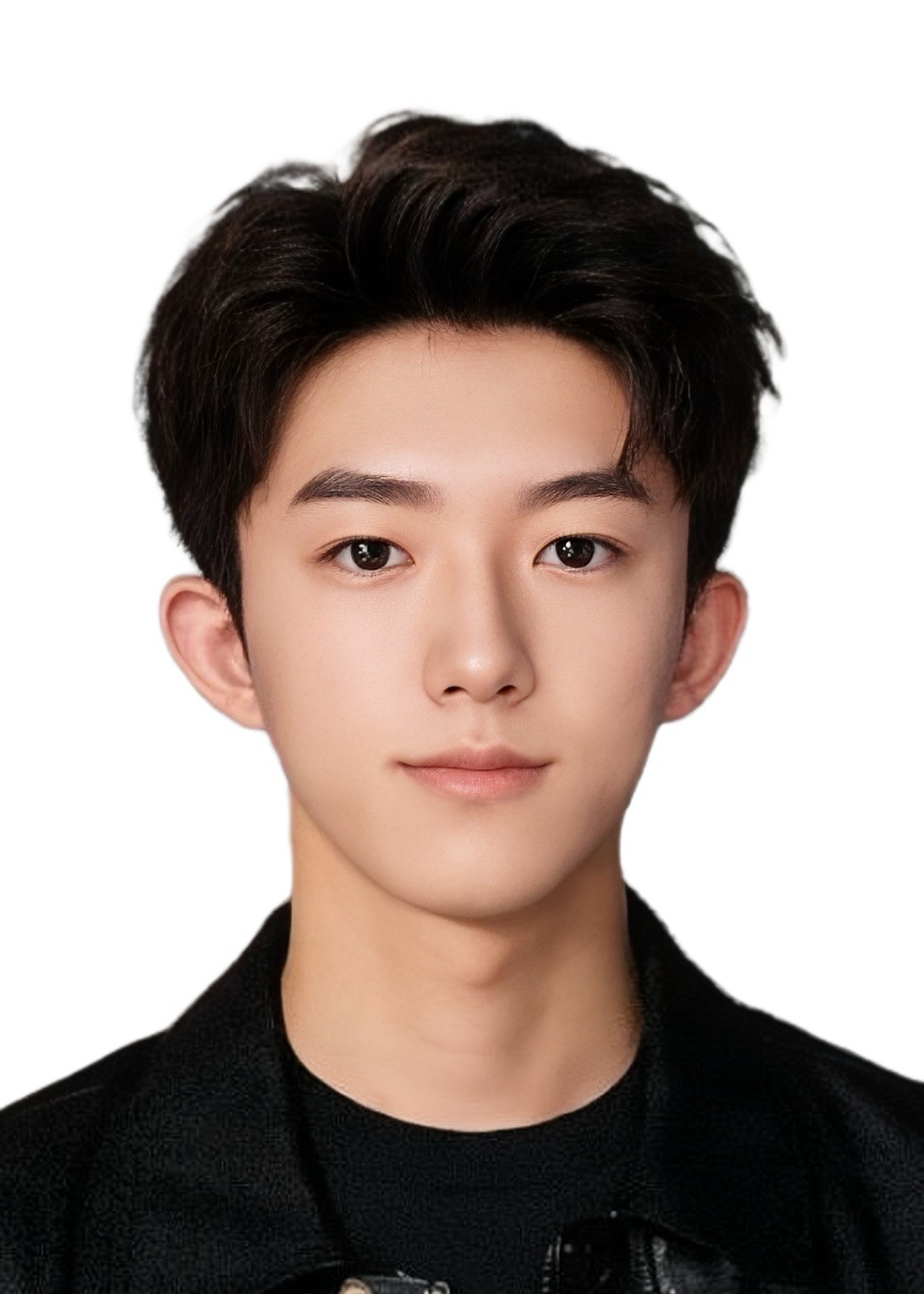} 
Zhenning Li received his Ph.D. in Civil Engineering from the University of Hawaii at Manoa in 2019. Currently, he holds the position of Assistant Professor at the State Key Laboratory of Internet of Things for Smart City, as well as the Departments of Civil and Environmental Engineering and Computer and Information Science at the University of Macau, Macau SAR.  He has published over 50 papers in top journals and conferences. His main areas of research focus on the intersection of connected autonomous vehicles and Big Data applications in urban transportation systems. He has been honored with several awards, including the TRB Best Young Researcher award and the CICTP Best Paper Award, amongst others.
\endbio

\end{document}